\documentclass[10pt]{article} 
\usepackage[accepted]{tmlr}

\usepackage{hyperref}
\usepackage{url}

\usepackage[utf8]{inputenc} 
\usepackage[T1]{fontenc}    
\usepackage{hyperref}       
\usepackage{url}            
\usepackage{booktabs}       
\usepackage{amsfonts}       
\usepackage{nicefrac}       
\usepackage{microtype}      
\usepackage{xcolor}         

\usepackage{graphicx}
\usepackage{makecell}
\usepackage{algorithm,algorithmic}
\usepackage{paralist,amsmath, amssymb,bm}
\usepackage{multirow}
\usepackage{ntheorem}
\newtheorem{theorem}{Theorem}
\newtheorem{proposition}{Proposition}
\newtheorem{lemma}{Lemma}
\newtheorem*{proof}{Proof}

\newtheorem{definition}{Definition}

\usepackage{enumitem}
\usepackage{textcase}
\usepackage{booktabs}
\usepackage{fancybox}
\usepackage{stackengine}

\usepackage{subfig}
\usepackage[capitalize]{cleveref}

\def\cR{\mathcal{R}}

\def\cW{\mathcal{W}}
\def\cS{\mathcal{S}}

\def\cU{\mathcal{U}}

\def\cW{\mathcal{W}}

\def\E{\mathbb E}
\def\R{\mathbb R}

\def\P{\mathbb P}

\def\op{\prime}

\DeclareMathOperator*{\argmax}{arg\,max}

\title{Multi-Output Distributional Fairness  via Post-Processing}

\author{\name Gang Li \email gang-li@tamu.edu \\
      \addr Texas A\&M University
      \ANDD
      \name Qihang Lin \email qihang-lin@uiowa.edu \\
      \addr The University of Iowa
      \ANDD
      \name Ayush Ghosh \email ayushghosh70@gmail.com \\
      \addr The University of Iowa
      \ANDD
      \name Tianbao Yang \email tianbao-yang@tamu.edu \\
      \addr Texas A\&M University
      }

\def\month{03}  
\def\year{2025} 
\def\openreview{\url{https://openreview.net/forum?id=MJOKrHqiV1}} 

\begin{document}

\maketitle

\begin{abstract}
The post-processing approaches are becoming prominent techniques to enhance machine learning models' fairness because of their intuitiveness, low computational cost, and excellent scalability.  However, most existing post-processing methods are designed for task-specific fairness measures and are limited to single-output models. In this paper, we introduce a post-processing method for multi-output models, such as the ones used for multi-task/multi-class classification and representation learning, to enhance a model's distributional parity, a task-agnostic fairness measure. Existing methods for achieving distributional parity rely on the (inverse) cumulative density function of a model’s output, restricting their applicability to single-output models.  Extending previous works, we propose to employ optimal transport mappings to move a model's outputs across different groups towards their empirical Wasserstein barycenter. An approximation technique is applied to reduce the complexity of computing the exact barycenter and a kernel regression method is proposed to extend this process to out-of-sample data. Our empirical studies evaluate the proposed approach against various baselines on multi-task/multi-class classification and representation learning tasks, demonstrating the effectiveness of the proposed approach.\footnote{Code is available at: \url{https://github.com/GangLii/TAB}} 
\end{abstract}

\section{Introduction}

In machine learning, multi-output learning is a broadly defined domain~\citep{xu2019survey,liu2018metric}, where the goal is to simultaneously predict multiple outputs given an input, such as multi-label classification, multi-class classification, multi-target regression, etc. In contrast to conventional single-output learning like binary classification, multi-output learning is characterized by its multi-variate nature, whose outputs exhibit rich information for further handling. Multi-output learning is important for real-world decision-making where final decisions are made by considering and weighting multiple factors and criteria.
For example, when applied to college admission, the predicted multi-outputs can represent a prospective student's likelihoods of accepting the offer, needing financial aid, completing the degree,  finding a job at graduation,  etc. 
Those outputs are weighted to guide admission decisions though the weights may vary with colleges, majors and years~\citep{IHE}.

However, multi-output learning for decision-making faces the challenge of bias and fairness. There is plenty of evidence indicating the discriminatory impact of ML-based decision-making on individuals and groups~\citep{o2017weapons,datta2014automated,bolukbasi2016man,barocas2016big,raji2019actionable}, such as racial bias in assessing the risk of recidivism~\citep{flores2016false} and gender bias in job advertising~\citep{simonite2015probing}. To mitigate the bias in machine learning,  numerous fairness criteria and algorithms have been proposed~\citep{corbett2017algorithmic,barocas2023fairness}.
These methods introduce statistical constraints during training or post-process the predictions to ensure fair treatment in accordance with corresponding fairness notions such as Demographic Parity~\citep{calders2009building, chuang2021fair}, Equality of Odds or Equal Opportunity~\citep{hardt2016equality, awasthi2020equalized}, Strong Demographic Parity~\citep{agarwal2019fair,jiang2020wasserstein} and AUC fairness~\citep{vogel2021learning,yang2023minimax,yao2023stochastic}.
Nevertheless, almost all existing methods focus on ensuring fairness for binary classification or regression within the context of single-output models. Extending fairness to multi-output settings, such as multi-label/multi-class classification and representation learning, remains underexplored and non-trivial. 

\begin{figure}[!t]
\footnotesize
\captionsetup[subfigure]{labelformat=empty}
  \centering
  \subfloat[][]{\includegraphics[width=.30\textwidth]{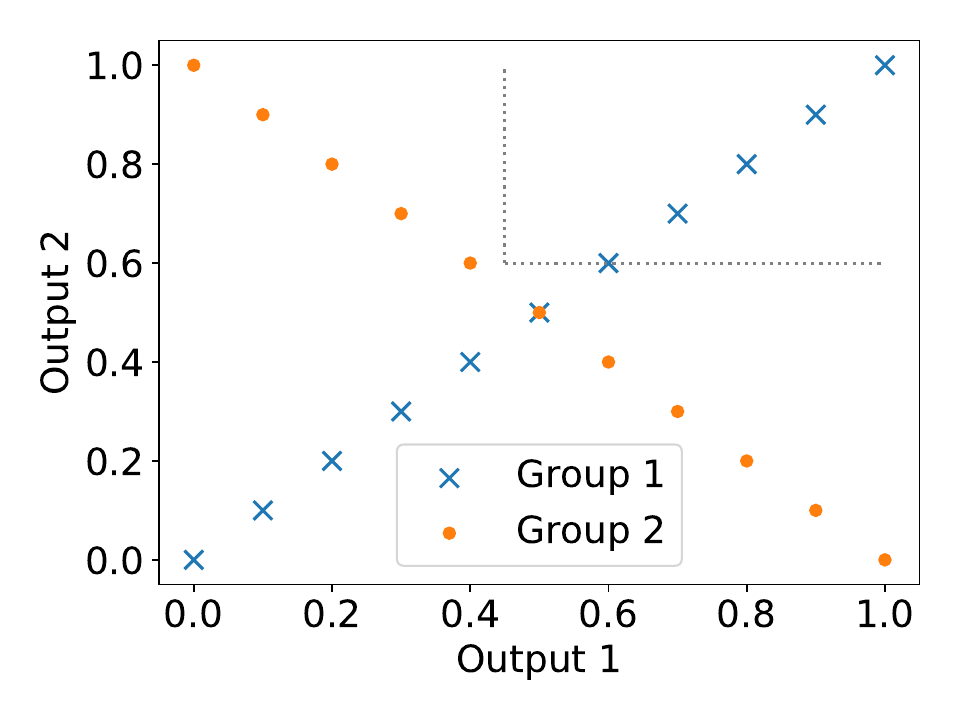}}
  \subfloat[][]{\includegraphics[width=.30\textwidth]{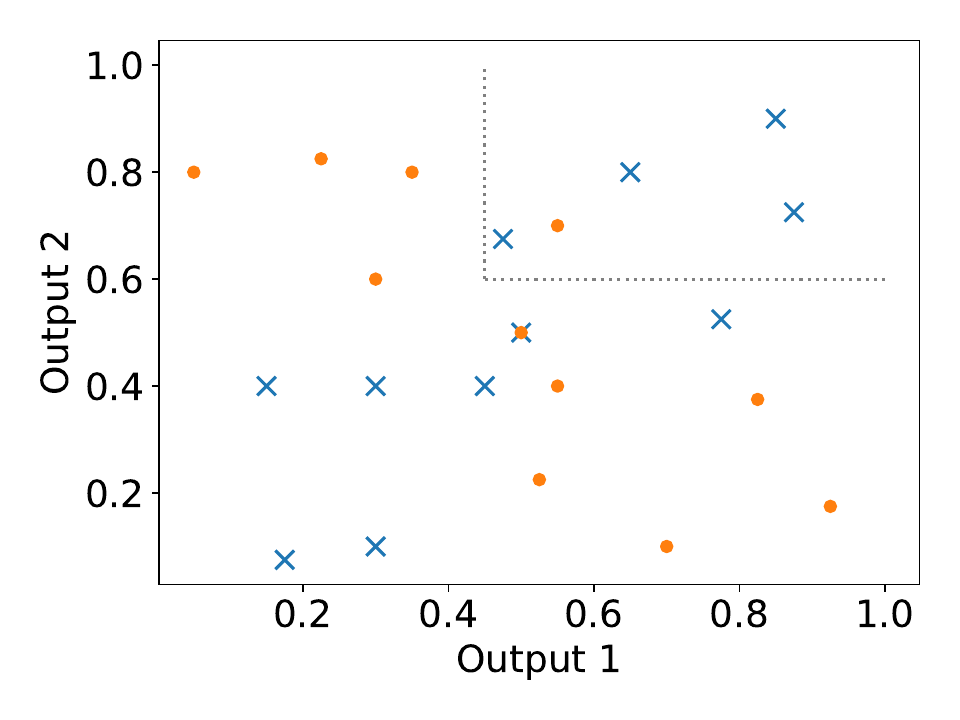}}
  \subfloat[][]{\includegraphics[width=.30\textwidth]{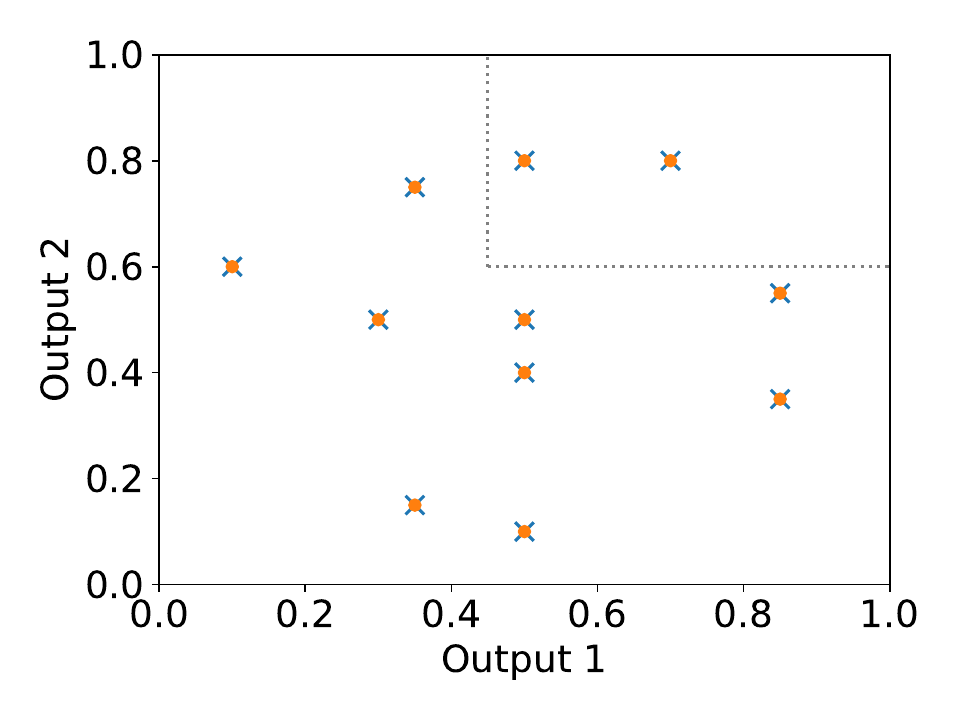}}
  \vspace*{-0.3in}
  \caption{Suppose the original model's outputs (left) minimize prediction error and have a good performance, one can balance performance and fairness by applying our method with setting $\alpha = 0.5$ (middle) or achieve exact fairness by setting  $\alpha = 0$ (right).}
  \label{fig:toyexample}\vspace*{-0.2in}
\end{figure}

A naive approach for multi-output fairness is to apply existing fairness-enhancing algorithms for single-output models to each output individually. However, removing unfairness in each output may not help reduce the unfairness in the joint distribution of the outputs. Consider a model with two outputs, with the distributions for two groups shown in Figure~\ref{fig:toyexample} (left). The outputs have the same marginal distribution in both groups, so each dimension of the outputs is considered to be fair. However, the outputs have very different joint distributions for different groups, leading to potential unfairness. For example, suppose Output1 and Output2 denote the likelihood of accepting the offer and completing the degree in a college admission procedure, then when a college requires outputs are higher than 0.45 and 0.6 respectively (shown in dot lines in Figure~\ref{fig:toyexample}), it leads to only students in group 1 being admitted. To address the challenge, in this paper, we propose a post-processing method to enhance fairness of multi-output models on multi-group data by producing a similar distribution of multi-dimensional outputs on each group with minimal manipulation(i.e., minimal mean squared error compared with original outputs).  Before discussing details, we present a 2D example in Figure~\ref{fig:toyexample}.

Measuring fairness based on the distribution of the multi-dimenstional outputs on each group leads to a strong notion called Statistical Parity or Strong Demographic Parity, which has been studied recently for single-output problems in literature~\citep{agarwal2019fair,jiang2020wasserstein,chzhen2020fair,hu2023fairness}. 
We apply the same notion to 
multi-output problems and call it Distributional Parity~(Definition~\ref{def:DP}). However, extending this kind of fairness from single-output problems to multi-output problems is non-trivial due to some technical and practical issues: (1) what is the optimal fair predictor for multi-output problems, which achieves Distributional Parity while preserving accuracy best; (2) empirically, how to compute the theoretical optimal fair predictor with only a set of observed data; (3) practically, how to generalize the post-processing method from training data to out-of-sample test data. Our approach addresses the mentioned challenges and our main contributions are summarized below:
\begin{itemize}[leftmargin=*]
    \item Following \citet{gouic2020projection,chzhen2020fair,chzhen2022minimax}, we derive a post-processing method by projecting a multi-output model to a constraint set defined by a fairness inequality that promotes distributional parity among multiple groups up to a user-specified tolerance of unfairness. We generalize the closed-form solution of the projection by \citet{chzhen2020fair,chzhen2022minimax} from single-output case to multi-output case using the optimal transport mappings to the Wasserstein barycenter of the model's outputs on each group.

    \item  The aforementioned closed form involves the density function of the model's outputs, which is unknown in practice. Although it can be approximated using training samples, the computation cost of the Wasserstein barycenter is still prohibitively high. To address this issue, we propose to replace the barycenter in the closed form by a low-cost approximate barycenter from \citet{lindheim2023simple} using training data. Additionally, we propose a kernel regression approach to extrapolate the closed-form solution, so the post-processing method can be applied to any out-of-sample data.

    \item The proposed method is task-agnostic and model-agnostic. Numerical experiments are conducted 
    to compare our method with various baselines on multi-label/multi-class classification and representation learning. The results demonstrate the effectiveness of our method.
\end{itemize}
 
\setlength{\abovedisplayskip}{2pt}
\setlength{\belowdisplayskip}{2pt}
\section{Related work}
The existing methods to achieve algorithmic fairness include three primary categories: 1) pre-processing methods that exclude sensitive features from data prior to training machine learning models~\citep{dwork2012fairness,du2018data,kamiran2012data,d2017conscientious}; 2) in-processing methods, which attain fairness during the training phase of the learning model~\citep{agarwal2018reductions,goh2016satisfying, zhang2018mitigating,kim2019learning,hong2021unbiased,du2021fairness,mo2021object,yao2023stochastic}; and 3) post-processing methods designed to alleviate unfairness in the model inferences after the learning process~\citep{hardt2016equality,lipton2018does,cui2023bipartite, petersen2021post,lohia2019bias,xian2023fair}. 
The pre-processing methods are inadequate if the sensitive information can be inferred from the remaining variables. The in-processing methods have high computational cost since they typically require re-training a model when the tolerance of unfairness changes. Our method belongs to the last category, which can be directly applicable to any pre-trained model and thus avoids the high computational cost from re-training a model. 

This work is motivated by \citet{gouic2020projection,chzhen2020fair,chzhen2022minimax} where task-agnostic post-processing methods are developed on the empirical cumulative density function and the quantile function of the output, which is limited to single-output models. We extend their approaches for multi-output models using the optimal transport mappings to the barycenter of the model's outputs on different groups. Different from their methods, we approximate the barycenter to reduce the computational cost and apply kernel regression to extend our post-processing method to any new data.  

There also exist post-processing methods for specific machine learning tasks. For example, assuming a Bayes optimal score function is available, the group thresholding methods by \citet{gaucher2023fair,chzhen2019leveraging,schreuder2021classification,zeng2022bayes} are developed for single-task binary classification. A similar thresholding method has been developed by \citet{denis2021fairness} for multi-class classification where the thresholds are computed using the Lagrangian multipliers of the demographic parity constraints. The method by \citet{xian2023fair} is also for multi-class classification where the output of the score function is mapped a class label by the optimal transport mapping to a Wasserstein barycenter with restricted supports. Compared to these works, the method in this paper is task-agnostic and can be applied to multi-task learning and (self-supervised) representation learning. 
\citet{hu2023fairness} developed an approach to enforce strong demographic parity in multi-task learning including both regression and binary classification tasks. Their method essentially applies the single-output method by \citet{gouic2020projection,chzhen2020fair} to each task separately, which may not guarantee the distributional parity in this paper. See the example in Figure~\ref{fig:toyexample}. Though there is a large number of literature working on learning representations independent of sensitive attributes, they achieve this by adversarial training~\citep{zhang2018mitigating,kim2019learning,xie2017controllable,madras2018learning}, variational auto-encoders~\citep{louizos2015variational,sarhan2020fairness,amini2019uncovering,gong2024towards}, or contrastive learning\citep{han2023dualfair,qi2024provable,zhang2022fairness,park2022fair}, which are mostly in-processing methods and thus have higher computational cost than our method in general.

The multi-output fairness problem in this paper is closely related to compositional fairness introduced in \citet{dwork2018fairness}, which highlights that individually fair classifiers may not compose into fair systems. While \citet{dwork2018fairness} focuses on categorical outputs, we examine continuous output distributions. Since categorical outputs often result from thresholding or softmax on continuous outputs, compositional fairness is a special case of multi-output fairness, and our method can directly address it.

\section{Preliminaries}

We consider a general multi-output machine learning problem, e.g., multi-output classification, multi-output regression and representation learning. Let $(X, S)$ be a random vector, where $X\in \R^d$ is the feature vector and $S\in \cS$ is a sensitive attribute. It is assumed that $S = [m] := \{1,...,m\}$.  A sample from the distribution of $(X, S)$ is denoted by $(x,s)$. We denote by $f^* : \R^d \times \cS \rightarrow \R^k$ a  machine learning model learned to generate a $k$-dimensional output. For example, $f^*$ can be a multi-task regression model to predict a vector of continuous values, a multi-class/multi-task classification model that returns the probability of each class/task, or a feature extractor that outputs a high-dimensional vector of an input data such as an image.

The focus of this work is not to train the model $f^*$.  Assuming $f^*$ is already obtained, this paper studies how to measure and improve its fairness using a post-processing method to modify the output $f^*(X,S)$. More specifically, given $f^*$, we consider a numerical approach for building a mapping $f : \R^d \times \cS \rightarrow \R^k$ that approximates $f^*$ but produces an output $f(X,S)$ more fair than $f^*(X,S)$. Here, the approximation error is measured by 
\begin{small}
 \begin{equation}
\label{eq:Rf}
\cR(f) = \E\|f^*(X,S)-f(X,S)\|_2^2.
\end{equation}   
\end{small}
If $f^*$ performs well for the task it was built for and the performance metric is continuous, a small $\cR(f)$ also ensures $f$ has a reasonably good performance, too.

\subsection{Distributional Parity}

Many existing definitions and measures of fairness in literature are task-specific including, for example, demographic parity, equalized odds and equal opportunity~\citep{dwork2012fairness,hardt2016equality}, which are specific to classification problems. In order to be task-agnostic and model-agnostic, we are interested in the fairness measure that is applied to the distribution of $f(X,S)$ and $f^*(X,S)$. A measure of this kind is the strong demographic parity introduced by \citet{agarwal2019fair,jiang2020wasserstein,chzhen2020fair}, which essentially indicates that a model $f(x,s)$ is fair only when $f(X,S)$ has the same distribution conditioning on $S=s$ for any $s\in \cS$. Although they only consider a single-output model, their fairness measure also applies to the multi-output case. Next we present their fairness measure formally with the name multi-output distributional parity. This new name helps differentiate it from the traditional demographic parity for binary classification~\citep{calders2009building}.

\begin{definition}\label{def:DP}
(Multi-output Distributional Parity) A measurable mapping $f : \R^d \times \cS \rightarrow \R^k$ satisfies distributional parity if, for every $s,s^\prime \in \cS$,
$f(X,S)$ conditioning on $S=s$ and $f(X,S)$ conditioning on $S=s'$  have the same distribution. 
\end{definition}

\textbf{Remark: }
Although we focus on 
Definition~\ref{def:DP}, an extension of demographic parity, it is natural to extend other notions of fairness, such as equal odds and equal opportunity, in a similar fashion. In particular, suppose $f(X,S)$ is used to predict a target variable $Y$ which takes values from a finite set $\mathcal{Y}$ (e.g., multi-class/mult-label classification). We say $f$ satisfies \textbf{distributionally equal opportunity} with respect to a class $y\in\mathcal{Y}$ if, for every $s,s^\prime \in \cS$, $f(X,S)$ conditioning on $(S,Y)=(s,y)$ and $f(X,S)$ conditioning on $(S,Y)=(s',y)$ have the same distribution. Similarly,  we say $f$ satisfies \textbf{distributionally equal odds} if, for every $s,s^\prime \in \cS$ and every $y\in\mathcal{Y}$, $f(X,S)$ conditioning on $(S,Y)=(s,y)$ and $f(X,S)$ conditioning on $(S,Y)=(s',y)$ have the same distribution.

Consider a joint distribution of $(X,S,Y)$ where $Y\in\mathbb{R}^p$ is a target vector. In the setting of a single-task binary classification, we have $k=1$, $p=1$ and $Y\in\{1,-1\}$ is the true class label. Then the model $f(X,S)\in\mathbb{R}$ typically represents a score/probability of positivity, which is used to predict $Y$ by comparing $f(X,S)$ with a threshold $\theta$, namely, the predicted label $\widehat Y$ generated as $\widehat Y=1$ if $f(X,S) \geq \theta$ and $\widehat Y=-1$ if $f(X,S)<\theta$. The traditional demographic parity~\citep{calders2009building,dwork2012fairness} holds if 
$
\P(\widehat Y=1|S=s)=\P(\widehat Y=1|S=s^\prime)\quad\forall s,s^\prime \in \cS.
$
Obviously, when $f$ satisfies distributional parity, it satisfies demographic parity for any threshold $\theta$. In fact, this is still true for multi-task binary classification problems, where $p=k>1$ and $Y\in\{1,-1\}^k$.

The notion of demographic parity has been extended for the multi-class classification problem, where $k>1$, $p=1$ and $Y\in[k]$~\citep{xian2023fair}. In this case, each coordinate of $f(X,S)$ represents the score of one class in $[k]$ and the predicted label is typically generated as $\widehat Y=\argmax_{l\in[k]}f_l(X,S)$. According to \citet{xian2023fair}, a model satisfies demographic parity when
\begin{small}
    \begin{align}\label{eqn:class_dp}
    \P(\widehat Y=y|S=s)=\P(\widehat Y=y|S=s^\prime)\quad\forall s,s^\prime \in \cS,~y\in[k],   
    \end{align}
\end{small}
which is clearly implied by distributional parity.

\subsection{Wasserstein Distance and Wasserstein Barycenters of Discrete Distributions}

It is challenging and, sometimes, unnecessary to obtain a model $f$ that exactly satisfies the distributional parity in Definition~\ref{def:DP}. In practice, a model slightly violating Definition~\ref{def:DP} can be acceptable for some applications. To quantify the extent to which a model $f$ satisfies the distributional parity, a statistical distance is often introduced to measure the difference between the distributions of $f(X,s)$ and $f(X,s^\prime)$ for any $s$ and $s^\prime$ in $\cS$. Following the literature~\citep{chzhen2020fair,gouic2020projection}, we utilize the Wasserstein distance that conveys the distinction between probability measures by quantifying the expense associated with transforming one probability measure into another.

\begin{definition}\label{def:w2}
(Wasserstein-2 distance). Let $\mu$ and $\nu$ be two probability measures on $\R^k$ with finite second moments. The squared Wasserstein-2 distance between $\mu$ and $\nu$ is defined as  
\begin{small}
\begin{align}
    \cW_2^2(\mu,\nu) = \inf_{\gamma \in \Gamma_{\mu,\nu}} \int_{\R^k \times \R^k}\|x-y\|^2_2 d\gamma(x,y)
\label{eqn:wasser}
\end{align}
\end{small}
where $\Gamma_{\mu,\nu}$ is the set of probability measures on $\R^k \times \R^k$ such that its marginal distributions are equal to $\mu$ and $\nu$, i.e.,  for all $\gamma \in \Gamma_{\mu,\nu}$ and all measurable sets $A,B \subset \R^k$ it holds that $\gamma(A\times \R^k) = \mu(A)$ and $\gamma(\R^k \times B) = \nu(B)$.
\end{definition}

Suppose probability measures $\mu$ and $\nu$ both have density functions and the infimum in \eqref{eqn:wasser} is obtained at $\gamma^*$. There exists a mapping $T_{\mu,\nu}: \R^k \rightarrow \R^k$ such that $\gamma^* = (\text{Id},T_{\mu,\nu})\#\mu$, where \# denotes the pushforward operator on measures and Id denotes the identity mapping~\citep{santambrogio2015optimal}. Here, $T_{\mu,\nu}$ is known as the \textbf{optimal transport mapping} from $\mu$ to $\nu$. With the Wasserstein distance, we can characterize the geometric average of finitely many probability distributions by the Wasserstein-2 barycenter, which will be later used in the measure of unfairness considered in this paper.  
\begin{definition}
\label{def:barycenter}
 For probability measures $\nu_1,\hdots, \nu_m$ and $p_1, \hdots, p_m$ such that $p_i>0, \sum_{i=1}^m p_i = 1$, the weighted Wasserstein-2 barycenter is given by
 \small
\begin{align}\label{eqn:barycenter}
   \nu^* = \arg\min_{\nu}\bigg\{\Psi(\nu):= \sum_{i=1}^m p_i\cW^2_2(\nu_i,\nu) \bigg\} 
\end{align}   
\normalsize
\end{definition}

\subsection{Post-processing with Distributional Parity Constraint}

Let $p_s = \P(S=s)$ and $\nu_{f|s}$ be the probability distribution of $f(X,S)$ conditioning on $S=s$ for $s\in\cS$. Following \citet{chzhen2022minimax}, we then measure the unfairness of model $f$ by the sum of the weighted distances between $\nu_{f|s}$ to their weighted Wasserstein-2 barycenters, namely, 
\begin{small}
\begin{equation}
\label{eq:Uf}
\mathcal{U}(f) =\min_{\nu}\sum_{s\in \cS} p_s\cW^2_2(\nu_{f|s},\nu).
\end{equation}
\end{small}
As Wasserstein-2 distance serves as a distance metric on the space of probability distributions, $\cW_2^2(\nu_{f|s}, \nu_{f|s^\op}) = 0$ if and only if $\nu_{f|s} = \nu_{f|s^\op}$~\citep{kwegyir2023repairing,peyre2017computational}. Hence, $\cU(f) =0$ if and only if for any $s, s^\prime \in\cS,\nu_{f|s} = \nu_{f|s^\prime}$, satisfying the distributional parity. However, when some level of unfairness is allowed, we only need to ensure $\mathcal{U}(f)\leq \alpha \mathcal{U}(f^*)$, where $\alpha\in[0,1]$ represents the tolerance of unfairness in terms of the fraction of the unfairness of $f^*$. Like \citet{chzhen2022minimax,kwegyir2023repairing}, we consider the post-processing problem with distributional parity constraint as follows:
\begin{equation}
\label{eq:postprocessing}
\min_f \cR(f)~\text{ s.t. }~ \cU(f) \leq \alpha\cU(f^*),
\end{equation}
where $\mathcal{R}(f)$ is defined in~\eqref{eq:Rf}.

\section{Structure of Optimal Solution}
Problem \eqref{eq:postprocessing} has been thoroughly studied by \citet{chzhen2020fair,chzhen2022minimax} under the setting that $f$ is a single-output model. They provide an intuitive closed form of the optimal solution of 
\eqref{eq:postprocessing} for any $\alpha\in[0,1]$ using the cumulative density function (CDF) of $\nu_{f^*|s}$ for each $s$. As they only focus on a single-output model, we expand their results to the multi-output case with some modifications in their proofs. We present the extension of their results in this section and highlight the main difference. 


When $\alpha=0$ in \eqref{eq:postprocessing}, the optimal solution of \eqref{eq:postprocessing} can be characterized by the following theorem. 

\begin{theorem}
\label{thm:char}
Suppose $\nu_{f^*|s}$ has density and finite second moments for each $s \in \cS$. Then
\begin{small}
\begin{align}
\label{eqn:char}
\min_{\mathcal{U}(f)=0}\mathcal{R}(f) =\min_\nu \sum_{s \in \cS} p_s \cW^2_2(\nu_{f^*|s}, \nu) = \mathcal{U}(f^*).
\end{align}    
\end{small}
Moreover, if $f_0$ and $\nu_0$ solve the first and second minimization in \eqref{eqn:char}, respectively, then $\nu_0$ is the distribution of $f_0$ and
\begin{small}
\begin{align}
\label{eqn:g_gene}
f_0(x,s) =T_{f^*|s,\nu^0}(f^*(x,s))
\end{align}    
\end{small}  
where $T_{f^*|s,\nu_0}:\R^k \rightarrow \R^k$ is the optimal transport mapping from $\nu_{f^*|s}$ to $\nu_0$.
\end{theorem}
By Definition~\ref{def:barycenter}, $\nu_0$ solving the second minimization in \eqref{eqn:char} is the barycenter of $\{\nu_{f^*|s}\}_{s\in\cS}$. This suggests a post-processing method by transporting the output $f^*(X,S)$ when $S=s$ to that barycenter. 

When $k=1$, Theorem~\ref{thm:char} is reduced to the structural results obtained by \citet{chzhen2020fair,gouic2020projection} (see, e.g., Theorem 2.3 in \citet{chzhen2020fair}). Indeed, when $k=1$, 
\begin{small}
\begin{equation}
\label{eq:TinoneD}
T_{f^*|s,\nu_0}(f^*(x,s))=\sum_{s^\prime \in S} p_{s^\prime} Q_{f^*|s^\prime}\circ F_{f^*|s}(f^*(x,s)),
\end{equation}
\end{small}
where $F_{f^*|s}$ is the CDF of $\nu_{f^*|s}$ and $Q_{f^*|s}$ is the quantile function of $\nu_{f^*|s}$, i.e., $Q_{f^*|s}(t)= \inf\{y\in\R: F_{f^*|s}(y)\geq t\}$.

In practice, when some level of unfairness is acceptable, one can select $\alpha>0$ in \eqref{eq:postprocessing}. When $\alpha=1$, $f=f^*$ is the optimal solution. When $\alpha\in (0,1)$, a closed form of the optimal solution for \eqref{eq:postprocessing} is derived by \citet{chzhen2022minimax,kwegyir2023repairing}. Although they are originally studied only under the single-output case, the same result holds for a multi-output model by almost the same proof. This closed form is presented in the following proposition, which motivates the way of trading off the error and fairness in our post-processing method in the next section. 

\begin{proposition}\label{lem:weight_ave}
(Proposition 4.1 in \citet{chzhen2022minimax}) Suppose $\nu_{f^*|s}$ has density and finite second moments for each $s \in \cS$. For any $\alpha \in [0,1]$, the optimal solution to \eqref{eq:postprocessing}, denoted by $f_\alpha$, satisfies (up to a zero-measure set)
\begin{align}
\label{eq:falpha}
    f_\alpha(x,s) &= \sqrt{\alpha}f^*(x,s) + (1-\sqrt{\alpha})f_0(x,s),
\end{align}
where $f_0$ is defined in Eq.~\eqref{eqn:g_gene}. 
\end{proposition}

\textbf{Remark:} \emph{It is not always possible to minimize $\cR$ and $\cU$ simultaneously by the same $f$. Therefore, a model $f$ is valuable as long as it achieves a value of $\cR$ (or $\cU$) that cannot be further reduced without increasing $\cU$ (or $\cR$). In fact, a  function $f(x,s)$ measurable in $x$ is called \emph{Pareto efficient} if there is no function $f'(x,s)$ measurable in $x$ such that one of the following two cases happens: (1) 
$\cR(f)\leq R(f')$ and $\cU(f)< U(f')$ or (2) $\cR(f)< R(f')$ and $\cU(f)\leq U(f')$. The Pareto frontier is 2-dimensional curve consisting of $(\cU(f),\cR(f))$ for all Pareto efficient $f$'s. Proposition 4.6 in \citet{chzhen2022minimax} shows, the Pareto frontier for $\cU$ and $\cR$ is actually the curve $\{(\cR(f_\alpha),\cU(f_\alpha))\}$ with $\alpha$ varying from 0 to 1, where $f_\alpha$ is from \eqref{eq:falpha}. Although $d=1$ in  \citet{chzhen2022minimax}, their Proposition 4.6 holds for our  setting also for any dimension $d\geq1$  by the same proof.}


\section{Fair Post-Processing with Finite Samples}

Theorem~\ref{thm:char} and Proposition~\ref{lem:weight_ave} only characterize the optimal solution to \eqref{eq:postprocessing} when $\nu_{f^*|s}$ has density. However, in practice, we only have access to a finite set of outputs of $f^*$, denoted by  $\{f^*(x_i,s_i)\}_{i=1}^n$, where $\{(x_i,s_i)\}_{i=1}^n$ is a dataset sampled from the distribution of $(X,S)$.~\footnote{The technique we propose is model-agnostic and can be applied directly to the outputs $\{f^*(x_i,s_i)\}_{i=1}^n$, so it does not require knowing $\{(x_i,s_i)\}_{i=1}^n$.} As a result, we are not able to compute $\nu_0$ and $T_{f^*|s,\nu_0}$ exactly and apply \eqref{eq:falpha}. To address this issue when $k=1$, a plug-in principle is applied by replacing $p_s$, $F_{f^*|s}(t)$, and $Q_{f^*|s}(t)$ in \eqref{eq:TinoneD} using their empirical approximation based on finite data sample~\citep{chzhen2020fair,gouic2020projection}. Because $f^*$ is a multi-output mapping in our case,  \eqref{eq:TinoneD} is not applicable. In the following, we present how to employ the plug-in principle by approximating $\nu_0$ and $T_{f^*|s,\nu_0}$ in \eqref{eqn:g_gene} by finite data sample. It consists of three main steps.

\subsection{Optimal Transportation with Finite Samples }

Consider approximating $\nu_0$ and $T_{f^*|s,\nu_0}$ by a collection of datasets $D_s = \{(x^s_i, s )\}^{n_s}_{i=1}$ for $s\in\cS$. We denote the empirical distribution of $f^*$ on $D_s$ by $\nu_{f^*(D_s)}$, i.e., 
$\nu_{f^*(D_s)}= \frac{1}{n_s}\sum_{i=1}^{n_s}\delta_{ f^*(x_i^s,s)} \text{ for }s\in\cS$,    
where $\delta$ is the Dirac measure. Consider two discrete distributions in $\mathbb{R}^k$:  
$\mu=\sum_{i=1}^{n_\mu}p_i^\mu\delta_{\xi_i^\mu}$
and $\nu=\sum_{i=1}^{n_\nu}p_i^\nu\delta_{\xi_i^\nu}$,
where $\xi_i^\mu\in\mathbb{R}^k$, $\xi_i^\nu\in\mathbb{R}^k$, $p_i^\mu>0$, $p_i^\nu>0$, $\sum_{i=1}^{n_\mu}p_i^\mu=1$ and $\sum_{i=1}^{n_\nu}p_i^\nu=1$. 
As a special case of \eqref{eqn:wasser}, the squared Wasserstein-2 distance between $\mu$ and $\nu$ is 
\small
\begin{eqnarray}\label{eqn:empOT}
           \cW_2^2(\mu,\nu) = \min\limits_{\gamma\in\R_+^{n_\mu \times n_\nu}} \sum_{i=1}^{n_\mu}\sum_{j=1}^{n_\nu} c_{ij}\gamma_{ij} \quad
         \text{ s.t. } \sum_{i=1}^{n_\nu} \gamma_{ij} = p_i^\mu ,\forall j,~~\sum_{j=1}^{n_\mu} \gamma_{ij} =p_i^\nu, \forall i. 
\end{eqnarray}
\normalsize
where $c_{ij} = \|\xi_i^\mu- \xi_j^\nu\|_2^2$ and $\gamma_{ij}$ represents the mass located $\xi_i^\mu$ transported to $\xi_j^\nu$ in order to move distribution $\mu$ to $\nu$. Also, $\gamma$ can be viewed as a discrete distribution in $\mathbb{R}^k\times\mathbb{R}^k$ supported on $(\xi_i^\mu,\xi_j^\nu)$ for $j =1, ..., n_\mu$ and $ i =1, ..., n_\nu$. Suppose $\gamma^*\in\R_+^{n_\mu \times n_\nu}$ is the optimal solution of \eqref{eqn:empOT}. The optimal transportation from $\mu$ to $\nu$ and from $\nu$ to $\mu$, denoted by $T_{\mu,\nu}$ and $T_{\nu,\mu}$ respectively, are random mappings such that 
\small
\begin{eqnarray}
\label{eq:discreteT}
\mathbb{P}(T_{\mu,\nu}(\xi_i^\mu) = \xi_j^\nu) = \frac{\gamma_{ij}^*}{p_i^\mu},\quad
\mathbb{P}(T_{\nu,\mu}(\xi_j^\nu) = \xi_i^\mu) = \frac{\gamma_{ij}^*}{p_j^\nu},
\end{eqnarray}
\normalsize
for $j =1, ..., n_\mu$ and $ i =1, ..., n_\nu$.

With $\cW_2^2(\mu,\nu)$ given in \eqref{eqn:empOT}, the barycenter of discrete distributions $\nu_i$ for $i=1,\dots,m$ is also defined as the solution of  \eqref{eqn:barycenter}. Since $\nu_{f^*(D_s)}$ is the discrete approximation of $\nu_{f^*|s}$, we propose to approximate $\nu_0$ in Theorem~\ref{thm:char} by the barycenter of $\{\nu_{f^*|s}\}_{s\in\cS}$ with the weights $p_s=\frac{n_s}{\sum_{s'\in\cS}n_{s'}}$. Unfortunately, Wasserstein barycenters are NP-hard to compute~\citep{altschuler2022wasserstein}. Although \eqref{eqn:barycenter} can be formulated as a multi-marginal optimal transport  problem~\citep{agueh2011barycenters} and solved as a linear program \citep{anderes2016discrete}, its computational complexity scales exponentially in terms of $m$. To reduce the exponential computing complexity, instead of computing the barycenter exactly, we adopt the approach by \citet{lindheim2023simple} to construct its approximation. This approach achieves a good balance between runtime and approximation error in practice. We present this approach in the next section.

\subsection{Approximate Barycenter with Finite Samples}

The approach by \citet{lindheim2023simple} first computes the optimal transport mapping between $\nu_{f^*(D_s)}$ to $\nu_{f^*(D_s')}$ for each pair of $s$ and $s'$ in $\cS$, i.e., $T_{\nu_{f^*(D_s)},\nu_{f^*(D_{s'})}}$ satisfying \eqref{eq:discreteT}. For simplicity of notation, we denote $T_{\nu_{f^*(D_s)},\nu_{f^*(D_{s'})}}$ by $T_{s,s'}$. Then, they define a mapping 
$M(f^*(x_i^s,s)) = \sum_{s'\in\cS}p_{s'} \mathbb{E}T_{s,s'}(f^*(x_i^s,s))$,
where $p_s=n_s/(\sum_{s'\in\cS}n_{s'})$ and the expectation is taken over the random output of $T_{s,s'}$ following distribution in \eqref{eq:discreteT}. Here, $M(f^*(x_i^s,s))$ is the weight average of the expected outcomes after transporting $f^*(x_i^s,s)$ to each of the $|\cS|$ distributions. 
Finally, the approximate barycenter is constructed as a discrete distribution with $\sum_{s\in\cS}n_s$ supports defined as follows
$\tilde\nu_{0}=\sum_{s\in\cS}\sum_{i=1}^{n_s}\frac{p_s}{n_s}\delta_{M(f^*(x_i^s,s))}$.
Compared to the exact barycenter, this approximation requires solving $|\cS|(|\cS|-1)/2$  optimal transport mapping between two discrete distributions and thus has a polynomial time complexity. This procedure is formally stated in Algorithm~\ref{alg:barycenter}.

\begin{algorithm}[tb]
   \caption{Approximate Barycenter}
   \label{alg:barycenter}
\begin{algorithmic}[1]
   \STATE {\bfseries Input:} A mapping $f^*: \R^d \times S \rightarrow \R^k$ and a dataset $D=\{(x_i, s_i)\}_{i=1}^n$ sampled from the distribution of $(X,S)$.
   \STATE Partition $D$ into subsets based on $s$ and obtain $D_s = \{(x^s_i, s )\}^{n_s}_{i=1}$ for $s\in\cS$.
   \STATE Let $p_s=\frac{n_s}{\sum_{s'\in\cS}n_{s'}}$ for $s\in\cS$.
   \FOR {$1\leq s < s'\leq |\cS|$}
        \STATE Solve \eqref{eqn:empOT} with $\mu=\nu_{f^*(D_s)}$ and $\nu=\nu_{f^*(D_{s'})}$ to obtain $T_{s,s'}=T_{\nu_{f^*(D_s)},\nu_{f^*(D_{s'})}}$.
   \ENDFOR
   \STATE Construct mapping $M(f^*(x_i^s,s)) = \sum_{s'\in\cS}p_{s'} \mathbb{E}T_{s,s'}(f^*(x_i^s,s))$
   \STATE {\bfseries Output:} Discrete distribution $\tilde \nu_0=\sum_{s\in\cS}\sum_{i=1}^{n_s}\frac{p_s}{n_s}\delta_{M(f^*(x_i^s,s))}$.
\end{algorithmic}
\end{algorithm}

Let $m=|\cS|$ and $\nu_s=\nu_{f^*(D_s)}$ in \eqref{eqn:barycenter}. Let $\hat \nu_0$ be optimal solution of \eqref{eqn:barycenter}, i.e, the barycenter of $\{\nu_{f^*(D_s)}\}_{s\in\cS}$. It is shown by \citet{lindheim2023simple} that $\Psi(\tilde \nu_0)/\Psi(\hat \nu_0)\leq 2$. 
This bound is significant because there is no polynomial-time algorithm can achieve a ratio arbitrarily close to one with high probability~\citep{altschuler2022wasserstein}.

Although Proposition~\ref{lem:weight_ave} is derived for continuous distribution, it motivates a heuristic post-processing method to update the output $f^*(x_i^s,s)$ to
\small
\begin{equation}
\label{eq:training_processed}
\tilde f_\alpha(x_i^s,s):=\sqrt{\alpha}f^*(x_i^s,s) + (1-\sqrt{\alpha}) T_{\nu_{f^*(D_s)}, \tilde \nu_0} (f^*(x_i^s,s))
\end{equation}
\normalsize
for $i=1,\dots,n_s$ and $s\in\cS$, where $\alpha\in[0,1]$. Note that $T_{\nu_{f^*(D_s)}, \tilde \nu_0}$ is not obtained during Algorithm~\ref{alg:barycenter} and needs to be solved separately. When $\alpha=0$ and $(x_i^s,s)$ is uniformly randomly sampled from $D_s$, $\tilde f_\alpha(x_i^s,s)$ has the same distribution for any $s$, indicating the post-processed outputs satisfy distributional parity. However, \eqref{eq:training_processed} is only defined for existing data in $D_s$ for $s\in\cS$. In the next section, we propose a method to extend this processing scheme to  new  samples from $(X,S)$.

\subsection{Post-Process Out-of-Sample Data}
Note that $\tilde f_\alpha(x_i^s,s)$ is defined for $(x_i^s,s)\in D_s$ only because the optimal transport mapping $ T_{\nu_{f^*(D_s)}, \tilde \nu_0}$ in  \eqref{eq:training_processed} is only defined on $f^*(x_i^s,s)$ with $(x_i^s,s)\in D_s$. To extend the definition of $\tilde f_\alpha(x_i^s,s)$ from $D_s$ to any $(x,s)\in\mathbb{R}^d\times \cS$, we extrapolate $T_{\nu_{f^*(D_s)}, \tilde \nu_0}$ over $\mathbb{R}^k$ using the Nadaraya-Watson kernel regression method~\citep{nadaraya1964estimating}. 

Let $K:\mathbb{R}^k\rightarrow\mathbb{R}_+$ be a kernel function satisfying $K(z)=K(-z)$and $\int_{\R^k} K(z)dz=1$. Let $h>0$ be a bandwidth. For any  $(x,s)\in\mathbb{R}^d\times \cS$, the kernel regression estimator of $T_{\nu_{f^*(D_s)}, \tilde \nu_0}(f^*(x,s))$ is 
\small
\begin{eqnarray}
\label{eq:test_T}
\bar f(x,s):=\tilde T_{\nu_{f^*(D_s)}, \tilde \nu_0}(f^*(x,s)):=\sum_{i=1}^{n_s}\frac{K\left((f^*(x,s)-f^*(x_i^s,s))/h\right) T_{\nu_{f^*(D_s)}, \tilde \nu_0}(f^*(x_i^s,s))
}{\sum_{j=1}^{n_s} K\left((f^*(x,s)-f^*(x_j^s,s))/h\right)}.
\end{eqnarray}
\normalsize
Here, $\bar f(x,s)$ is the post-processed prediction for data $(x,s)$. Recall that $T_{\nu_{f^*(D_s)}, \tilde \nu_0}$ is a randomly mapping (see \eqref{eq:discreteT}), so is  $\tilde T_{\nu_{f^*(D_s)}, \tilde \nu_0}$. 
To analyze  the fairness of $\bar f(x,s)$ on the out-of-sample data distribution, we present the following proposition to characterize how much $\bar f(X,S)$ violates the distributional parity after post-processing by bounding 
$\cU(\bar f(X,S))$ and the Wasserstein distance between the distributions of $\bar f(X,S)$ on different groups. The proof is given in Section~\ref{sec:prooflemma}.

\setlength{\abovedisplayskip}{2pt}
\setlength{\belowdisplayskip}{2pt}

\begin{proposition}\label{lem:kernel_map}
Suppose $K(z)=\frac{1}{\sqrt{2\pi}}\exp(-z^2)$. Conditioning on data $D=\{(x_i, s_i)\}_{i=1}^n$, it holds that
\begin{equation}
\label{eq:kernel_map_error}
\cU(\bar f(X,S))\leq \sum_{s\in\cS}O\left(p_sn_s^2\exp\left(-\frac{1}{h^2}\right)\right)+\sum_{s\in\cS}O\left(\frac{p_s}{h^4}\cW^2_2(\nu_{f^*|s},\nu_{f^*(D_s)})\right).
\end{equation}
Moreover, let $\nu_{\bar f|s}$ be the distribution of $\bar f(X,S)$ conditioning on $S=s$ for $s\in\cS$. It holds that
\begin{equation}
\label{eq:kernel_map_error_betweengroups}
\cW^2_2(\nu_{\bar f|s},\nu_{\bar f|s'})\leq  O\left((n_s^2+n_{s'}^2)\exp\left(-\frac{1}{h^2}\right)\right)+ O\left(\frac{1}{h^4}\cW^2_2(\nu_{f^*|s},\nu_{f^*(D_s)})\right).
\end{equation}
\end{proposition}
In both upper bounds above, the first term is the error due to extrapolation which will converge to zero as the bandwidth $h$ approaches zero. The second term is the error due to the sampling error between training data $D$ and the ground-truth data distribution. The second error is represented by the Wasserstein distribution $W^2_2(\nu_{f^*|s},\nu_{f^*(D_s)})$. As the data sizes $n_s$ for $s\in\cS$ increase, the  rate at which  $W^2_2(\nu_{f^*|s},\nu_{f^*(D_s)})$ converges to zero
has been well studied in literature. For example, according to case (3) of Theorem 2  in \cite{fournier2015rate}, if there exists $q>4$ such that 
$$
\mathbb{E}_{(X,S)|S=s}\|f^*(X,S)\|_2^q<+\infty,
$$
where $\mathbb{E}_{(X,S)|S=s}$ is the expectation over $(X,S)$ conditioning on $S=s$, then the Wasserstein distance in 
the second term in \eqref{eq:kernel_map_error} and \eqref{eq:kernel_map_error_betweengroups} satisfies the following concentration inequality
\begin{align}
\label{eq:wdistance_highprob}
\mathbb{P}(W^2_2(\nu_{f^*|s},\nu_{f^*(D_s)})\geq \eta)\leq a(n_s,\eta)+b(n_s,\eta)\text{ for any }\eta\in(0,1)
\end{align}
where
$$
a(n_s, \eta) = C \begin{cases} 
\exp(-c n_s \eta^2), & \text{if } k<4, \\ 
\exp\big(-c n_s \big(\eta / \log(2 + 1/\eta)\big)^2\big), & \text{if } k=4, \\ 
\exp(-c n_s \eta^{k/2}), & \text{if } k>4.
\end{cases}\quad
\text{ and }\quad
b(n_s, \eta)=n_s^{-\frac{q-4}{4}}\eta^{-\frac{q}{4}}.
$$


After extrapolating $T_{\nu_{f^*(D_s)}, \tilde \nu_0}$ to $\tilde T_{\nu_{f^*(D_s)}, \tilde \nu_0}$ defined above, we can extend \eqref{eq:training_processed} for any out-of-sample data by updating $f^*(x,s)$ to 
\small
\begin{equation}
\label{eq:testing_processed}
\tilde f_\alpha(x,s):=\sqrt{\alpha}f^*(x,s) + (1-\sqrt{\alpha})\bar f(x,s).
\end{equation}
\normalsize
We then formally present this post-processing method in Algorithm~\ref{alg:post_process}. Note that when $(x,s)\in D_s$, we still apply the original mapping in \eqref{eq:training_processed} instead of its extrapolation \eqref{eq:testing_processed}.

Our post-processing method can also be extended to improve the fairness of a model under the notion of distributionally equal opportunity or distributionally equal odds introduced in the remark after Definition~\ref{def:DP}. In particular, to improve the fairness in terms of distributionally equal odds, we can partition the training set $D$ into $|\mathcal{Y}|$ subsets $\{D^y\}_{y\in\mathcal{Y}}$ depending on the class label $Y$. Then we apply Algorithm~\ref{alg:post_process} to each subset $D^y$ to construct a separate post-processing mapping $\bar f_y(x,s)$ like \eqref{eq:test_T} for each combination of group $s$ and class $y$. To apply it on an out-of-sample data point, we first generate a predicted label $\widehat{y}$ using $f^*(x,s)$ and then convert $f^*(x,s)$ to $\bar f_{\widehat{y}}(x,s)$ using the mapping constructed on  $D^y$. The same procedure works for distributionally equal opportunity with respect to one class $y\in\mathcal{Y}$, except that we only need to construct the mapping on $D^y$ and post-process data for one class $y$.

\begin{algorithm}[tb]
   \caption{Post-Processing Method by Transporting to Approximate Barycenter (TAB)}
   \label{alg:post_process}
\begin{algorithmic}[1]
  \STATE {\bfseries Input:} A data point $(x,s)\in\mathbb{R}^d\times \cS$, a mapping $f^*: \R^d \times S \rightarrow \R^k$ and a dataset $D=\{(x_i, s_i)\}_{i=1}^n$ sampled from the distribution of $(X,S)$.
   \STATE Partition $D$ into subsets based on $s$ and obtain $D_s = \{(x^s_i, s )\}^{n_s}_{i=1}$ for $s\in\cS$.
   \STATE Compute $\tilde \nu_0$ by Algorithm~\ref{alg:barycenter}.
    \FOR {$s=1,\dots, |\cS|$}
        \STATE Solve \eqref{eqn:empOT} with $\mu=\nu_{f^*(D_s)}$ and $\nu=\tilde\nu_0$ to obtain $T_{\nu_{f^*(D_s)}, \tilde \nu_0}$.
   \ENDFOR
   \STATE
   \textbf{if} $(x,s)\in D_s$ \textbf{then}  compute $\tilde f_\alpha(x,s)$ as in \eqref{eq:training_processed} \textbf{else} compute $\tilde f_\alpha(x,s)$ as in \eqref{eq:testing_processed}.
   \STATE {\bfseries Output:} $\tilde f_\alpha(x,s)$
\end{algorithmic}
\end{algorithm}

\section{Experiments}
\label{sec:exp}
In this section, we apply the proposed post-processing method to machine learning models with multiple outputs to evaluate its effectiveness, including multi-label/multi-class classification and representation learning. 

\textbf{Datasets.} In our experiments, we include four datasets from various domains, including marketing domain(\textit{Customer} dataset \footnote{\url{https://www.kaggle.com/datasets/kaushiksuresh147/customer-segmentation}}),
medical diagnosis( \textit{Chexpert} Dataset~\citep{irvin2019chexpert}), face recognition (\textit{CelebA} dataset~\citep{liu2015faceattributes} and \textit{UTKFace} dataset~\citep{zhifei2017cvpr}). The  details of these datasets are provided in Appendix~\ref{sec:data}. 

\textbf{Baselines and Settings.} To verify the effectiveness, we compare our method against six baseline approaches, including four post-processing methods: (1)\textit{Hu et al. (2023)}~\citep{hu2023fairness}, which introduces a post-processing approach for incorporating fairness into multi-task learning using multi-marginal Wasserstein barycenters;  (2)\textit{Xian et al (2023)}~\citep{xian2023fair}, which characterizes the inherent tradeoff of demographic parity in multi-class classification and proposes an effective post-processing method;  (3)\textit{FRAPPE}~\citep{tifreafrappe}, which proposes a generic framework that converts any regularized in-processing method into a post-processing approach;
(4)\textit{Adv Debiasing}~\citep{zhang2018mitigating}, which presents a framework for mitigating demographic group biases by introducing a group-related variable and jointly learning a predictor and adversary, 
and two in-processing methods: (5)\textit{SimFair}~\citep{liu2023simfair}, which focuses on training models for fairness-aware multi-label classification; (6)\textit{f-FERM}~\citep{baharloueif}, which presents a unified stochastic optimization framework for fair empirical risk minimization regularized by f-divergence measures.

Specifically in our experiments, for baseline \textit{Adv Debiasing}, both Predictor Block and Adversary Block are implemented by a two-layer neural network with 128 hidden units and \textit{tanh} activating function, respectively. For baselines \textit{FRAPPE}, \textit{SimFair}, and \textit{Adv Debiasing}, we train the models for 60 epochs with Adam optimizer, batch size as 64, and tune the learning rate in \{1e-3, 1e-4\}.
We vary their weight parameter $\lambda$(or $\alpha$) in \{  0.1, 1, 2, 4, 8, 10\} to show the trade-off between classification performance and fairness. For \textit{f-FERM}, we follow their paper to vary the weight parameter $\lambda$ in \{0.1, 1, 10, 50, 100, 150\} and tune the learning rate in \{0.1, 0.01, 0.001\} with their proposed optimization algorithm.  For \textit{Hu et al. (2023)}~\citep{hu2023fairness}, we vary parameter $\alpha$ for their method in \{0, 0.2, 0.4, 0.6, 0.8, 1.0\}. Following their paper, the parameter $\alpha$ for \citet{xian2023fair} are in \{1, 0.16, 0.14, 0.12, 0.1, 0.08, 0.06, 0.04, 0.02, 0.01, 0.008, 0.006, 0.004, 0.002, 0.001, 0.0.\}. For our method,  we experiment with a Gaussian kernel and $h$ is chosen from \{0.02, 0.04, 0.5, 1\} based on input dimension, as smaller $h$ theoretically and empirically leads to better performance but too small $h$ may lead to numerical issues.  We vary $\alpha$ for our method in \{0, 0.2, 0.4, 0.6, 0.8, 1.0\}. All the experiments are run five times with different seeds.


\subsection{Multi-label classification}

\begin{figure}[!tb]  
  \centering
   \subfloat[][]{\includegraphics[width=.33\textwidth]{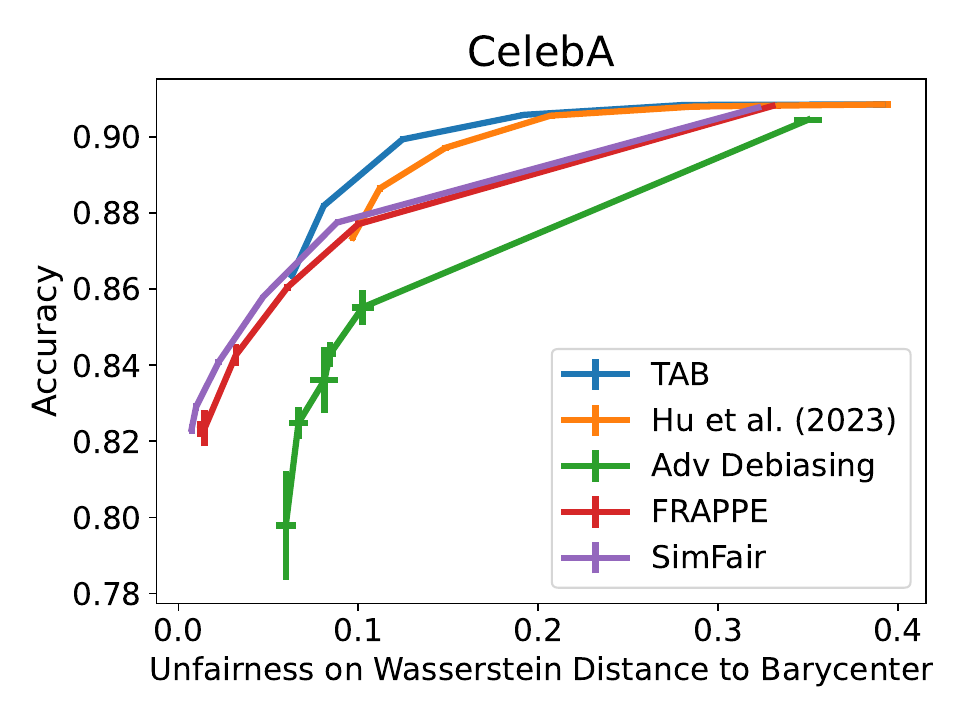}}
  \subfloat[][]{\includegraphics[width=.33\textwidth]{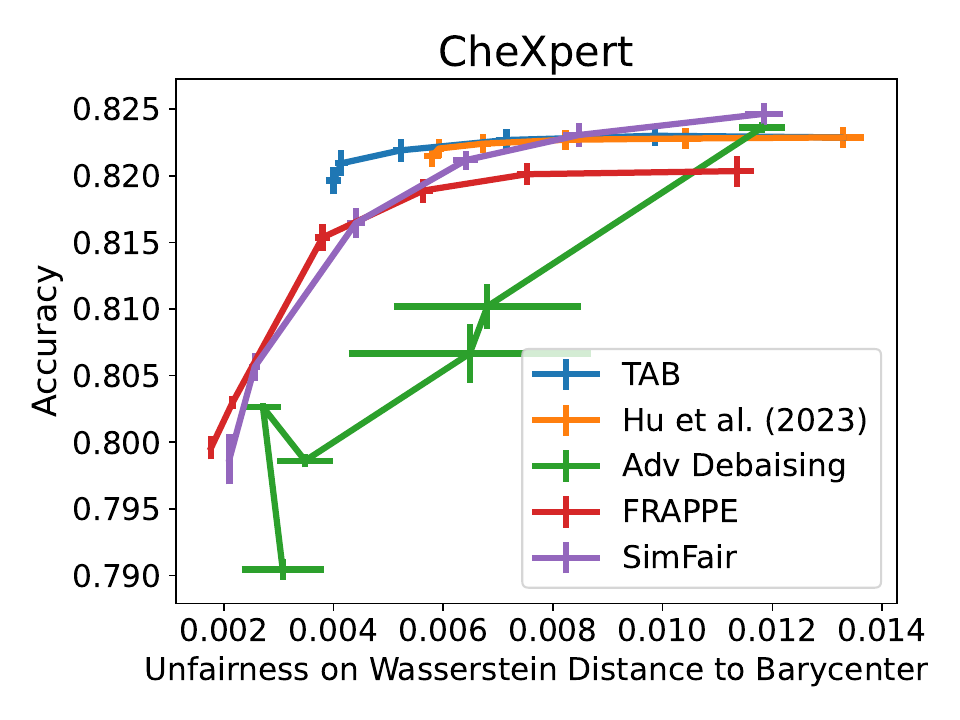}}
  \subfloat[][]{\includegraphics[width=.33\textwidth]{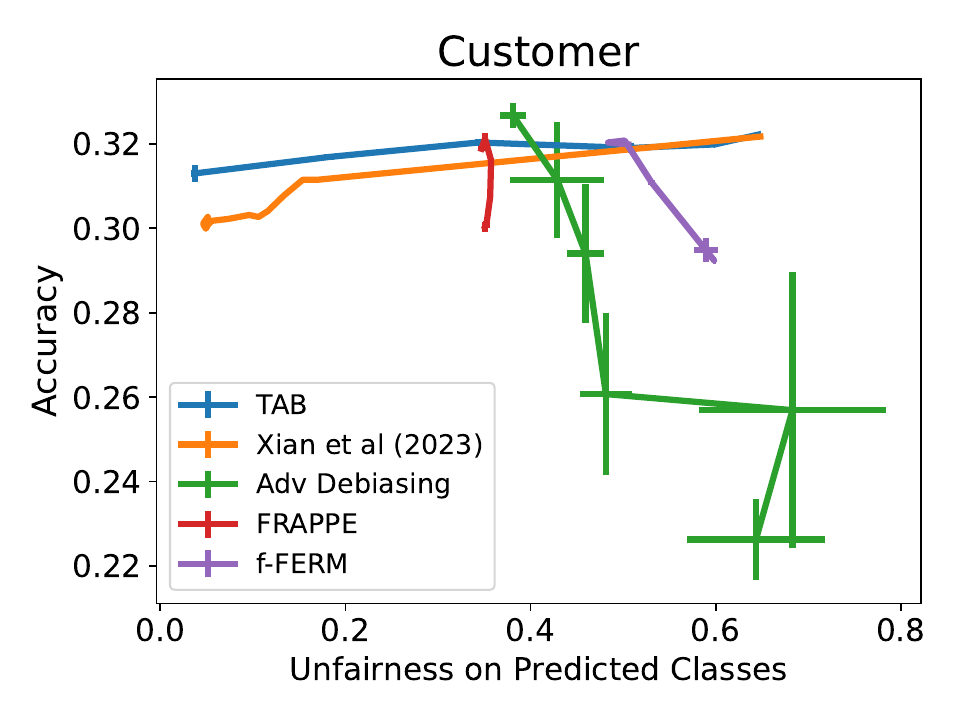}}
  \caption{Multi-label classification on CelebA dataset (a) and Chexpert dataset (b); (c) Multi-class classification on Customer dataset.}
  \label{fig:mclass}
\end{figure}

For multi-label classification tasks, we experiment on  \textit{CelebA} dataset and \textit{Chexpert} dataset. We compare our method with \textit{SimFair}~\citep{liu2023simfair}, \textit{FRAPPE}~\citep{tifreafrappe}, \textit{Adv Debiasing} and the method \textit{Hu et al. (2023)}~\citep{hu2023fairness}, which essentially independently applies the post-processing method for a single-output model in \citet{chzhen2020fair,chzhen2022minimax} to each coordinate of the output of a multi-task classification model. Firstly, one ResNet50~\citep{he2016deep} and one DenseNet121~\citep{huang2017densely} and are trained on the \textit{CelebA} and \textit{Chexpert} data respectively, then post-processing methods \textit{Hu et al. (2023)}~\citep{hu2023fairness}, \textit{Adv Debiasing} and our proposed method are applied to the predicted probabilities of each task and methods \textit{SimFair}~\citep{liu2023simfair} and \textit{FRAPPE}~\citep{tifreafrappe} are applied to the extracted image features to train a new linear classification head.  For both \textit{FRAPPE} and \textit{SimFair}, the fairness regularizer is defined as described in Equation 1 of the ~\citep{liu2023simfair} paper, as it has been demonstrated to be effective for promoting demographic parity. The results are summarized in Fig. \ref{fig:mclass}(a) and \ref{fig:mclass}(b). We can observe that, at the same level of accuracy, our method achieves much lower unfairness than baselines, which can be seen clearly when drawing horizontal lines across the figures. Particularly, on the Chexpert dataset, our method delivers fairer predictions with only a negligible decrease in predictive performance. From Fig. \ref{fig:mclass}(a) and \ref{fig:mclass}(b), we also note that learning-based baselines such as \textit{FRAPPE}, \textit{SimFair}, and \textit{Adv Debiasing} tend to achieve lower unfairness but only at the cost of a substantial drop in accuracy, whereas processing-based methods like \textit{Hu et al. (2023)} and our method maintain a better balance. Furthermore, we observe that with the same fairness regularizer, the post-processing method \textit{FRAPPE} achieves comparable fairness-error trade-offs to the in-processing technique \textit{SimFair}, aligning with the findings reported in ~\citet{tifreafrappe}.

\begin{figure*}[!tb] 
  \centering
  \subfloat[][]{\includegraphics[width=.24\textwidth]{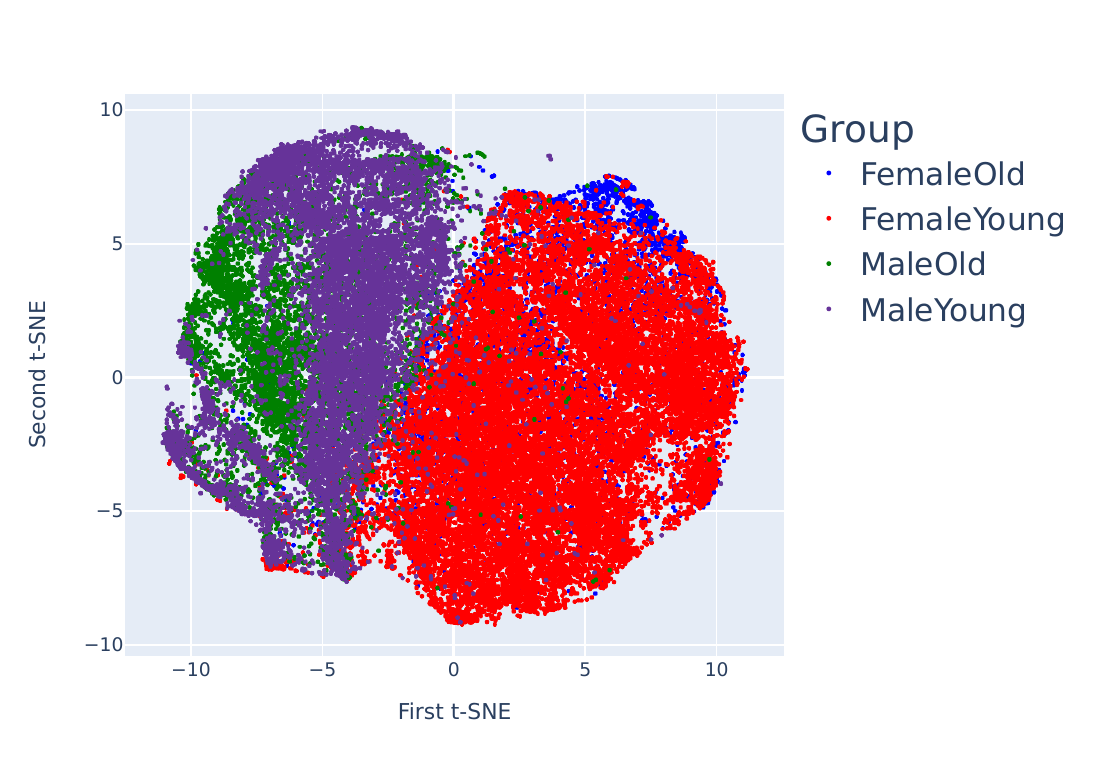}}
  \subfloat[][]{\includegraphics[width=.24\textwidth]{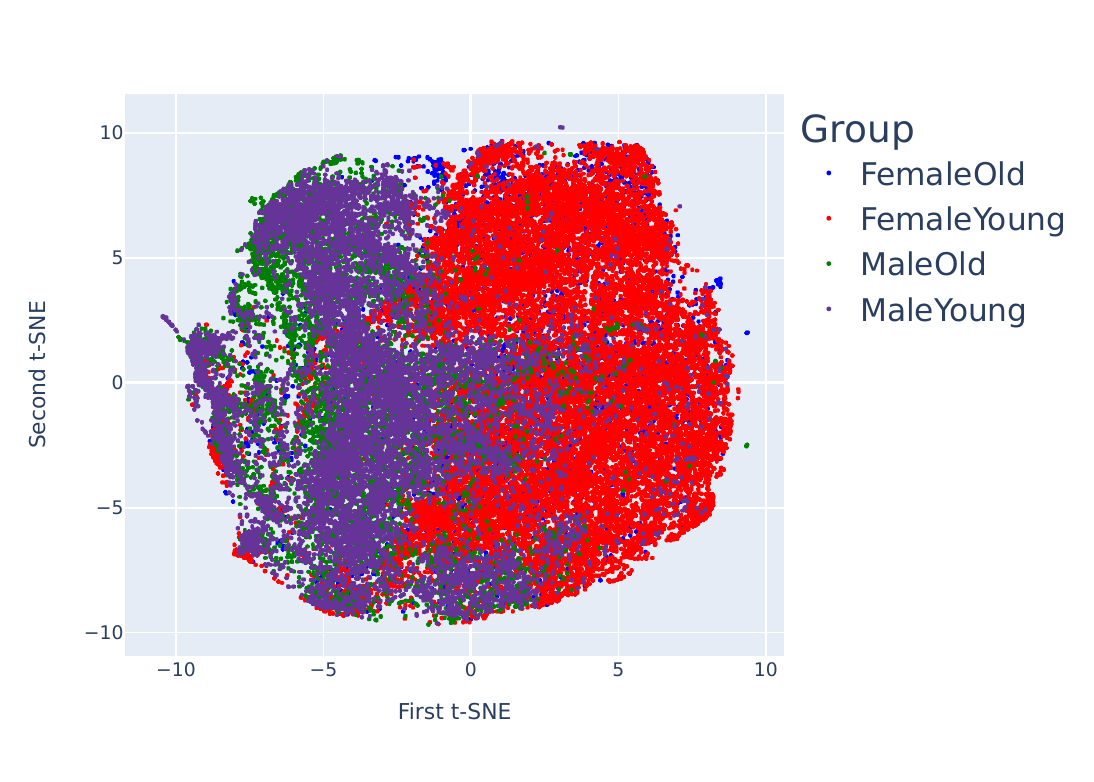}}
  \subfloat[][]{\includegraphics[width=.24\textwidth]{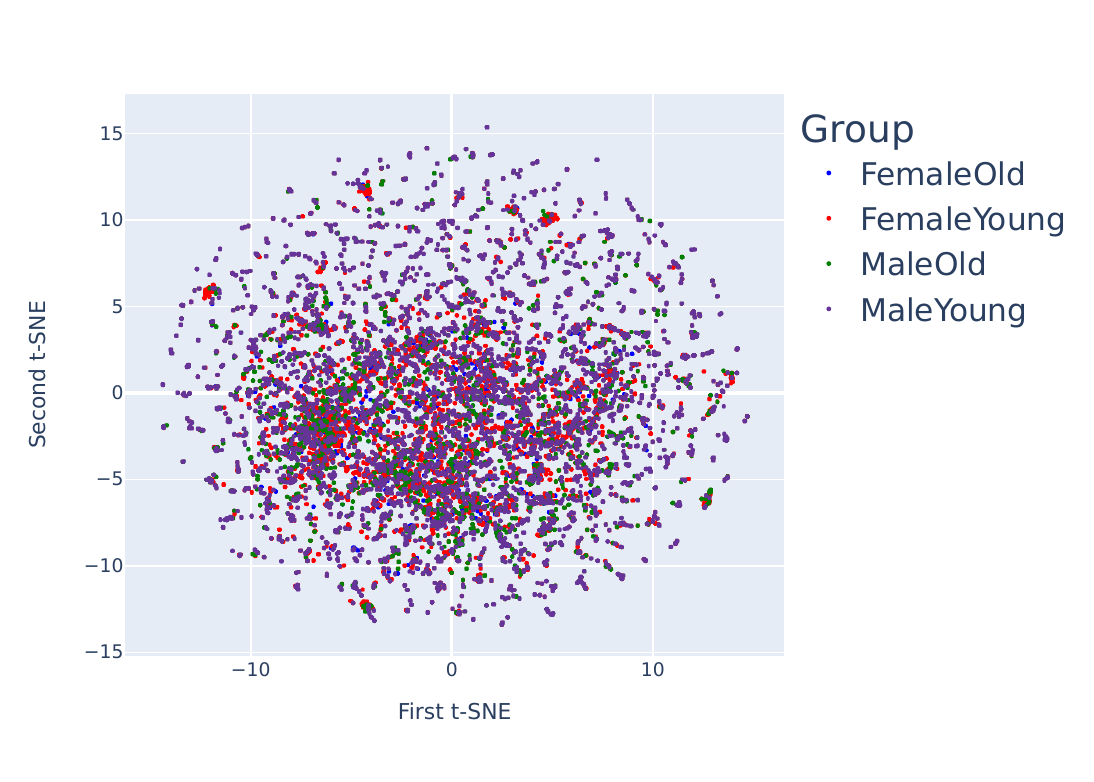}}
  \subfloat[][]{\includegraphics[width=.24\textwidth]{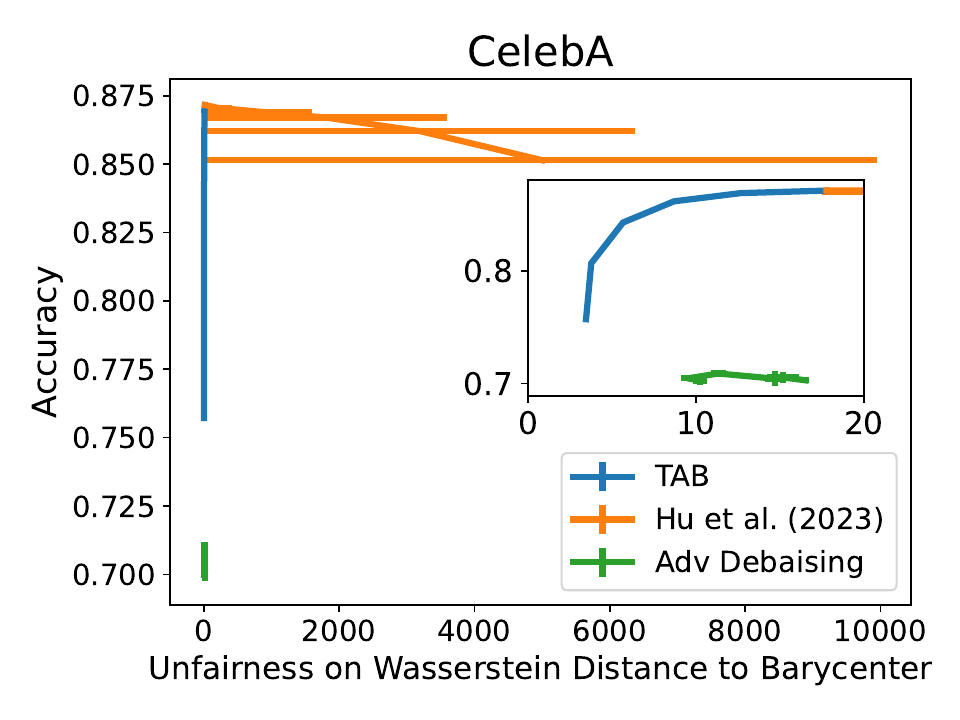}}

  \caption{t-SNE visualization of representations on CelebA dataset. (a) Raw representations from an SSL model; (b) Representations after post-processing($\alpha=0$) by \citet{hu2023fairness}; (c) Representations after post-processing($\alpha=0$) with TAB(Ours); (d) Performance on downstream tasks.}
  \label{fig:fea1} %
 
  \centering
  \subfloat[][]{\includegraphics[width=.24\textwidth]{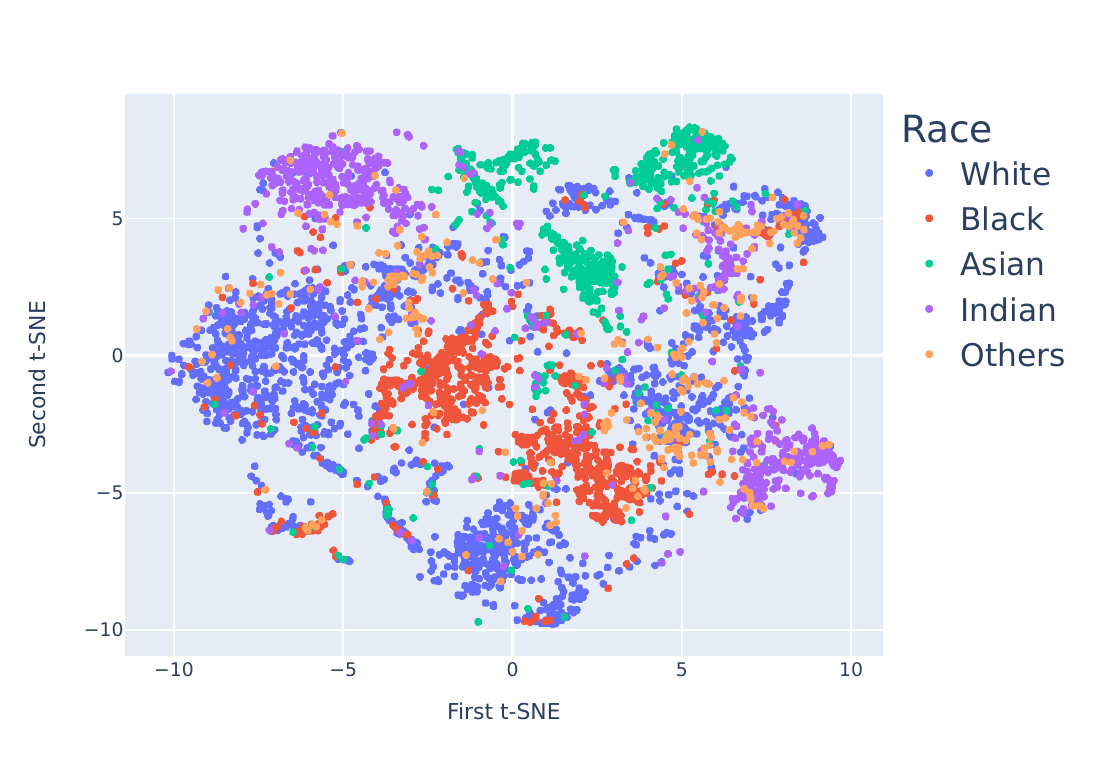}}
  \subfloat[][]{\includegraphics[width=.24\textwidth]{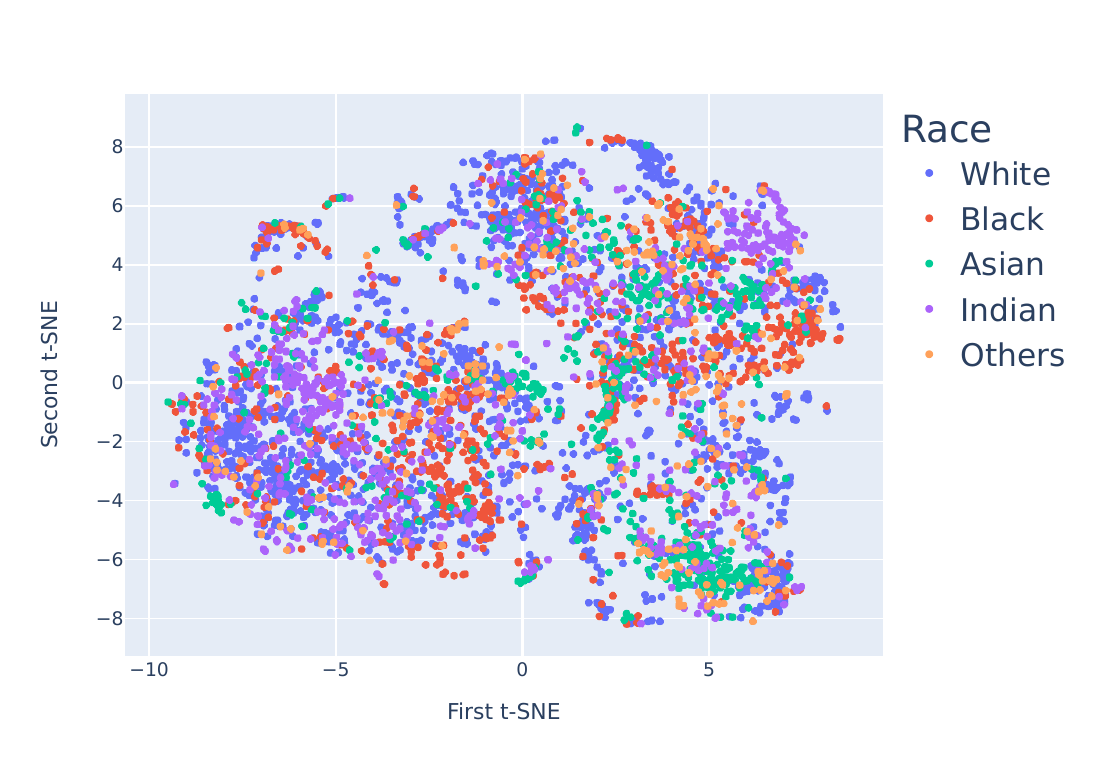}}
  \subfloat[][]{\includegraphics[width=.24\textwidth]{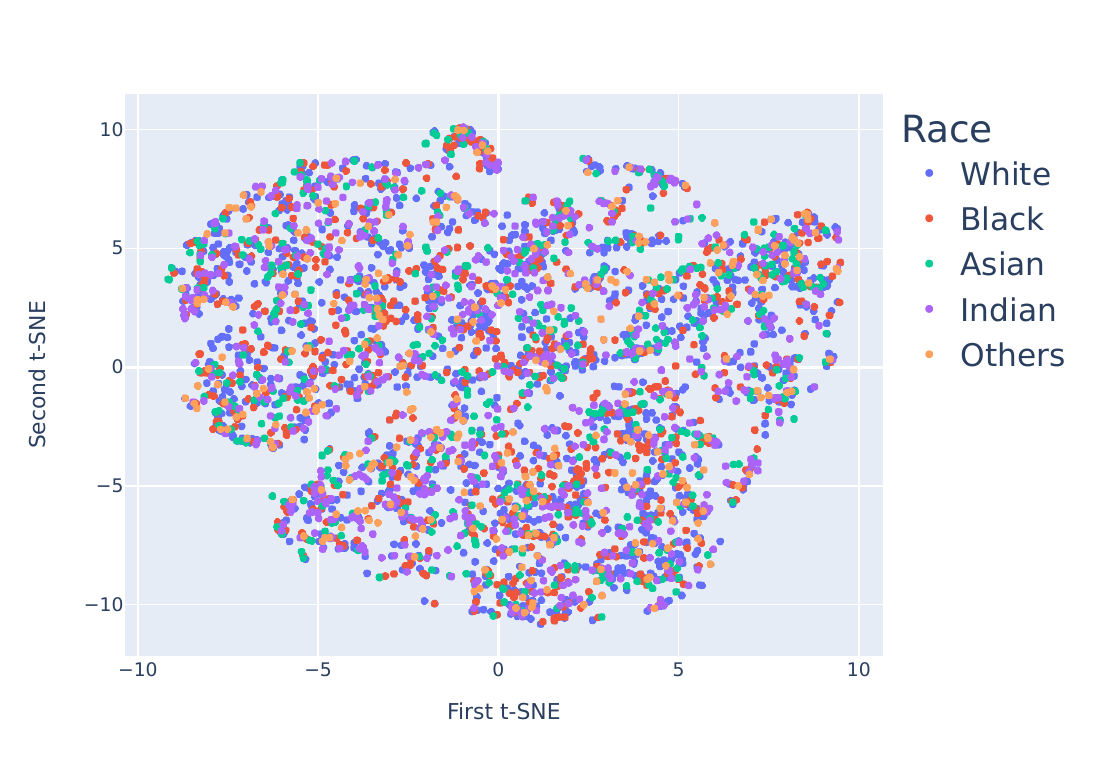}}
  \subfloat[][]{\includegraphics[width=.24\textwidth]{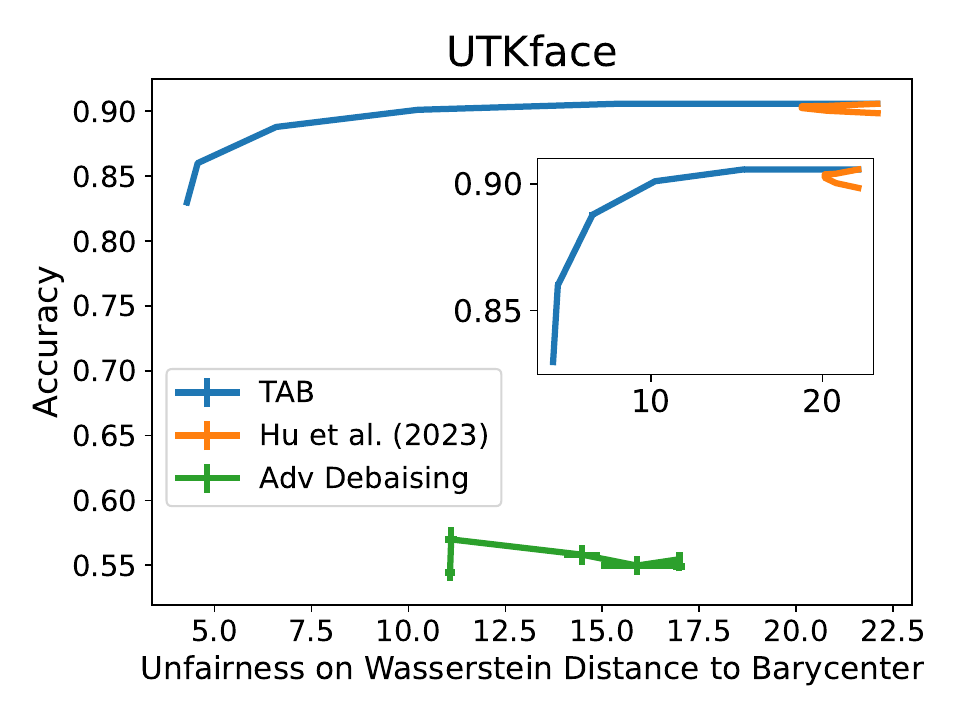}}

  \caption{t-SNE visualization of representations on UTKface Dataset. (a) Raw representations from a CLIP model; (b) Representations after post-processing($\alpha=0$) with \citet{hu2023fairness}; (c) Representations after post-processing($\alpha=0$) with TAB(Ours); (d) Performance on downstream tasks.}
  \label{fig:fea2}  
\end{figure*}

\subsection{Multi-class classification}

To verify the effectiveness of our proposed method on multi-class classification tasks, we apply our method to the \textit{customer} dataset.  A classic multi-class logistic regression model is built first on the training data and then post-processing methods are applied to the predicted probabilities of each class. For in-processing methods, the multi-class logistic regression models are trained with fairness regularizer from scratch. We compare our method with \textit{FRAPPE}~\citep{tifreafrappe}, \textit{f-FERM}~\citep{baharloueif}, \textit{Xian et al (2023)}~\citep{xian2023fair} and \textit{Adv Debiasing}.  Note that method \textit{Xian et al (2023)} directly produces the class labels instead of score function, so we can only perform the comparison on the classification accuracy and the unfairness based on Equation~\ref{eqn:class_dp}, which is in favor of their method. For \textit{FRAPPE}, the fairness regularizer is defined as the Equation~\ref{eqn:class_dp} proposed by \citet{xian2023fair}. We present the results in Fig. \ref{fig:mclass}(c). The figure demonstrates that our method effectively reduces unfairness measured by the metric proposed by \citet{xian2023fair} and achieves a superior balance between accuracy and unfairness compared to the baselines, particularly when a higher level of fairness is required. Besides, we notice that although it is implied that methods \textit{f-FERM}~\citep{baharloueif} and \textit{FRAPPE}~\citep{tifreafrappe} are applicable for multi-class classification, their respective papers do not include experiments for this scenario, and their methods, in our experiments, fail to mitigate unfairness defined by Equation~\ref{eqn:class_dp} even when it brings a significant undesirable performance decrease.

\subsection{Representation learning}
In this part, we explore the fair representation learning for self-supervised learning(SSL) models and large pre-trained foundation models. We experiment on \textit{CelebA} dataset and \textit{UTKFace} dataset. Due to the scarcity of the postprocessing methods for representation learning in existing literature, we still compare our method with the post-processing method \textit{Hu et al. (2023)}~\citep{hu2023fairness} and \textit{Adv Debiasing}. For CelebA dataset, we first train an SSL model to learn representations with a dimension of 128, by employing the algorithm proposed in \cite{yuan2022provable} on the whole training dataset. Then we apply post-processing methods to eliminate the sensitive information in the raw representations. Specifically, for \textit{Adv Debiasing}, we utilize the raw representations as the labels for the Predictor as there is no task-specific labels in self-supervised learning. In other words, we aim for the model to remove sensitive information while altering the raw representation as minimally as possible.
With processed representations, we also build a multi-label logistic regression model to evaluate the performance on downstream tasks. For UTKFace dataset, raw representations are generated by a released CLIP(ViT-B/16) model~\citep{radford2021learning} which is pre-trained on 400M text-image pairs, with a dimension of 512. 
 
 We evaluate the unfairness of the representations as well as the accuracy performance of the downstream tasks. The results are summarized in Fig.~\ref{fig:fea1} and \ref{fig:fea2}. By examining the representations of Fig.~\ref{fig:fea1}(a) and \ref{fig:fea2}(a), it's evident that both the SSL model and CLIP model disclose sensitive attributes prominently. Even after applying post-processing techniques proposed in \textit{Hu et al. (2023)}~\citep{hu2023fairness}, this exposure persists, as depicted in Fig.~\ref{fig:fea1}(b) and Fig.~\ref{fig:fea2}(b). However, through our proposed methods illustrated in Fig.~\ref{fig:fea1}(c) and Fig.~\ref{fig:fea2}(c), we are able to achieve fairer representations with respect to sensitive groups. Notably, from Fig.~\ref{fig:fea1}(d) and \ref{fig:fea2}(d), we can observe that our method achieves a better tradeoff between downstream performance and distributional parity, whereas \textit{Hu et al. (2023)}~\citep{hu2023fairness} fails to improve the distributional parity because, as discussed in Fig.~\ref{fig:toyexample}, the elimination of unfairness in individual outputs may not necessarily mitigate unfairness in the joint distribution of outputs.

\section{Conclusion and Discussion}
In this paper,  we have proposed a post-processing method to enhance fairness for multi-output machine learning models, which is underexplored in the literature.
Our approach employs optimal transport to move a model’s outputs across different groups towards their empirical Wasserstein barycenter to achieve the model’s distributional parity. We have developed an approximation technique to reduce the complexity of computing the exact barycenter and a kernel regression method for extending this process to out-of-sample data.  Extensive experimental results on multi-label/multi-class classification and representation learning demonstrate the effectiveness of our method.

One limitation of this work is that the notion of fairness in Definition~\ref{def:DP} we pursue is strong while a weak notion such as \eqref{eqn:class_dp} might be sufficient for some specific applications. To achieve a stronger sense of fairness may lead to more decreases in the predictive performance than targeting a weak sense. Therefore, applicability of the proposed method may vary depending on the specific use case. Additionally, the lack of theoretical convergence analysis of the proposed method as the training sample size increases is another limitation, which is an important future work.

\subsubsection*{Broader Impact Statement}

This paper is to attain distributional fairness in machine learning-driven decision-making, particularly concerning various demographic groups such as gender, age, and race. When enforcing group fairness, it may potentially disadvantage certain individuals. Other than this, none we feel must be specifically highlighted here.


\subsubsection*{Acknowledgments}
We thank anonymous reviewers for constructive comments. G. Li, Q. Lin, and T. Yang were partially supported by NSF  Award 2147253.

\bibliography{tmlr}

\begin{thebibliography}{78}
\providecommand{\natexlab}[1]{#1}
\providecommand{\url}[1]{\texttt{#1}}
\expandafter\ifx\csname urlstyle\endcsname\relax
  \providecommand{\doi}[1]{doi: #1}\else
  \providecommand{\doi}{doi: \begingroup \urlstyle{rm}\Url}\fi

\bibitem[Agarwal et~al.(2018)Agarwal, Beygelzimer, Dud{\'\i}k, Langford, and Wallach]{agarwal2018reductions}
Alekh Agarwal, Alina Beygelzimer, Miroslav Dud{\'\i}k, John Langford, and Hanna Wallach.
\newblock A reductions approach to fair classification.
\newblock In \emph{International conference on machine learning}, pp.\  60--69. PMLR, 2018.

\bibitem[Agarwal et~al.(2019)Agarwal, Dud{\'\i}k, and Wu]{agarwal2019fair}
Alekh Agarwal, Miroslav Dud{\'\i}k, and Zhiwei~Steven Wu.
\newblock Fair regression: Quantitative definitions and reduction-based algorithms.
\newblock In \emph{International Conference on Machine Learning}, pp.\  120--129. PMLR, 2019.

\bibitem[Agueh \& Carlier(2011)Agueh and Carlier]{agueh2011barycenters}
Martial Agueh and Guillaume Carlier.
\newblock Barycenters in the wasserstein space.
\newblock \emph{SIAM Journal on Mathematical Analysis}, 43\penalty0 (2):\penalty0 904--924, 2011.

\bibitem[Altschuler \& Boix-Adsera(2022)Altschuler and Boix-Adsera]{altschuler2022wasserstein}
Jason~M Altschuler and Enric Boix-Adsera.
\newblock Wasserstein barycenters are np-hard to compute.
\newblock \emph{SIAM Journal on Mathematics of Data Science}, 4\penalty0 (1):\penalty0 179--203, 2022.

\bibitem[Amini et~al.(2019)Amini, Soleimany, Schwarting, Bhatia, and Rus]{amini2019uncovering}
Alexander Amini, Ava~P Soleimany, Wilko Schwarting, Sangeeta~N Bhatia, and Daniela Rus.
\newblock Uncovering and mitigating algorithmic bias through learned latent structure.
\newblock In \emph{Proceedings of the 2019 AAAI/ACM Conference on AI, Ethics, and Society}, pp.\  289--295, 2019.

\bibitem[Anderes et~al.(2016)Anderes, Borgwardt, and Miller]{anderes2016discrete}
Ethan Anderes, Steffen Borgwardt, and Jacob Miller.
\newblock Discrete wasserstein barycenters: Optimal transport for discrete data.
\newblock \emph{Mathematical Methods of Operations Research}, 84:\penalty0 389--409, 2016.

\bibitem[Awasthi et~al.(2020)Awasthi, Kleindessner, and Morgenstern]{awasthi2020equalized}
Pranjal Awasthi, Matth{\"a}us Kleindessner, and Jamie Morgenstern.
\newblock Equalized odds postprocessing under imperfect group information.
\newblock In \emph{International conference on artificial intelligence and statistics}, pp.\  1770--1780. PMLR, 2020.

\bibitem[Baharlouei et~al.(2024)Baharlouei, Patel, and Razaviyayn]{baharloueif}
Sina Baharlouei, Shivam Patel, and Meisam Razaviyayn.
\newblock f-ferm: A scalable framework for robust fair empirical risk minimization.
\newblock In \emph{The Twelfth International Conference on Learning Representations}, 2024.

\bibitem[Barocas \& Selbst(2016)Barocas and Selbst]{barocas2016big}
Solon Barocas and Andrew~D Selbst.
\newblock Big data's disparate impact.
\newblock \emph{California law review}, pp.\  671--732, 2016.

\bibitem[Barocas et~al.(2023)Barocas, Hardt, and Narayanan]{barocas2023fairness}
Solon Barocas, Moritz Hardt, and Arvind Narayanan.
\newblock \emph{Fairness and machine learning: Limitations and opportunities}.
\newblock MIT Press, 2023.

\bibitem[Bolukbasi et~al.(2016)Bolukbasi, Chang, Zou, Saligrama, and Kalai]{bolukbasi2016man}
Tolga Bolukbasi, Kai-Wei Chang, James~Y Zou, Venkatesh Saligrama, and Adam~T Kalai.
\newblock Man is to computer programmer as woman is to homemaker? debiasing word embeddings.
\newblock \emph{Advances in neural information processing systems}, 29, 2016.

\bibitem[Calders et~al.(2009)Calders, Kamiran, and Pechenizkiy]{calders2009building}
Toon Calders, Faisal Kamiran, and Mykola Pechenizkiy.
\newblock Building classifiers with independency constraints.
\newblock In \emph{2009 IEEE international conference on data mining workshops}, pp.\  13--18. IEEE, 2009.

\bibitem[Chuang \& Mroueh(2021)Chuang and Mroueh]{chuang2021fair}
Ching-Yao Chuang and Youssef Mroueh.
\newblock Fair mixup: Fairness via interpolation.
\newblock \emph{arXiv preprint arXiv:2103.06503}, 2021.

\bibitem[Chzhen \& Schreuder(2022)Chzhen and Schreuder]{chzhen2022minimax}
Evgenii Chzhen and Nicolas Schreuder.
\newblock A minimax framework for quantifying risk-fairness trade-off in regression.
\newblock \emph{The Annals of Statistics}, 50\penalty0 (4):\penalty0 2416--2442, 2022.

\bibitem[Chzhen et~al.(2019)Chzhen, Denis, Hebiri, Oneto, and Pontil]{chzhen2019leveraging}
Evgenii Chzhen, Christophe Denis, Mohamed Hebiri, Luca Oneto, and Massimiliano Pontil.
\newblock Leveraging labeled and unlabeled data for consistent fair binary classification.
\newblock \emph{Advances in Neural Information Processing Systems}, 32, 2019.

\bibitem[Chzhen et~al.(2020)Chzhen, Denis, Hebiri, Oneto, and Pontil]{chzhen2020fair}
Evgenii Chzhen, Christophe Denis, Mohamed Hebiri, Luca Oneto, and Massimiliano Pontil.
\newblock Fair regression with wasserstein barycenters.
\newblock \emph{Advances in Neural Information Processing Systems}, 33:\penalty0 7321--7331, 2020.

\bibitem[Corbett-Davies et~al.(2017)Corbett-Davies, Pierson, Feller, Goel, and Huq]{corbett2017algorithmic}
Sam Corbett-Davies, Emma Pierson, Avi Feller, Sharad Goel, and Aziz Huq.
\newblock Algorithmic decision making and the cost of fairness.
\newblock In \emph{Proceedings of the 23rd acm sigkdd international conference on knowledge discovery and data mining}, pp.\  797--806, 2017.

\bibitem[Cui et~al.(2023)Cui, Pan, Zhang, and Wang]{cui2023bipartite}
Sen Cui, Weishen Pan, Changshui Zhang, and Fei Wang.
\newblock Bipartite ranking fairness through a model agnostic ordering adjustment.
\newblock \emph{IEEE Transactions on Pattern Analysis and Machine Intelligence}, 2023.

\bibitem[d'Alessandro et~al.(2017)d'Alessandro, O'Neil, and LaGatta]{d2017conscientious}
Brian d'Alessandro, Cathy O'Neil, and Tom LaGatta.
\newblock Conscientious classification: A data scientist's guide to discrimination-aware classification.
\newblock \emph{Big data}, 5\penalty0 (2):\penalty0 120--134, 2017.

\bibitem[Datta et~al.(2014)Datta, Tschantz, and Datta]{datta2014automated}
Amit Datta, Michael~Carl Tschantz, and Anupam Datta.
\newblock Automated experiments on ad privacy settings: A tale of opacity, choice, and discrimination.
\newblock \emph{arXiv preprint arXiv:1408.6491}, 2014.

\bibitem[Denis et~al.(2021)Denis, Elie, Hebiri, and Hu]{denis2021fairness}
Christophe Denis, Romuald Elie, Mohamed Hebiri, and Fran{\c{c}}ois Hu.
\newblock Fairness guarantee in multi-class classification.
\newblock \emph{arXiv preprint arXiv:2109.13642}, 2021.

\bibitem[Du et~al.(2021)Du, Mukherjee, Wang, Tang, Awadallah, and Hu]{du2021fairness}
Mengnan Du, Subhabrata Mukherjee, Guanchu Wang, Ruixiang Tang, Ahmed Awadallah, and Xia Hu.
\newblock Fairness via representation neutralization.
\newblock \emph{Advances in Neural Information Processing Systems}, 34:\penalty0 12091--12103, 2021.

\bibitem[du~Pin~Calmon et~al.(2018)du~Pin~Calmon, Wei, Vinzamuri, Ramamurthy, and Varshney]{du2018data}
Flavio du~Pin~Calmon, Dennis Wei, Bhanukiran Vinzamuri, Karthikeyan~Natesan Ramamurthy, and Kush~R Varshney.
\newblock Data pre-processing for discrimination prevention: Information-theoretic optimization and analysis.
\newblock \emph{IEEE Journal of Selected Topics in Signal Processing}, 12\penalty0 (5):\penalty0 1106--1119, 2018.

\bibitem[Dwork \& Ilvento(2018)Dwork and Ilvento]{dwork2018fairness}
Cynthia Dwork and Christina Ilvento.
\newblock Fairness under composition.
\newblock \emph{arXiv preprint arXiv:1806.06122}, 2018.

\bibitem[Dwork et~al.(2012)Dwork, Hardt, Pitassi, Reingold, and Zemel]{dwork2012fairness}
Cynthia Dwork, Moritz Hardt, Toniann Pitassi, Omer Reingold, and Richard Zemel.
\newblock Fairness through awareness.
\newblock In \emph{Proceedings of the 3rd innovations in theoretical computer science conference}, pp.\  214--226, 2012.

\bibitem[Edwards(2011)]{edwards2011kantorovich}
David~A Edwards.
\newblock On the kantorovich--rubinstein theorem.
\newblock \emph{Expositiones Mathematicae}, 29\penalty0 (4):\penalty0 387--398, 2011.

\bibitem[Flores et~al.(2016)Flores, Bechtel, and Lowenkamp]{flores2016false}
Anthony~W Flores, Kristin Bechtel, and Christopher~T Lowenkamp.
\newblock False positives, false negatives, and false analyses: A rejoinder to machine bias: There's software used across the country to predict future criminals. and it's biased against blacks.
\newblock \emph{Fed. Probation}, 80:\penalty0 38, 2016.

\bibitem[Fournier \& Guillin(2015)Fournier and Guillin]{fournier2015rate}
Nicolas Fournier and Arnaud Guillin.
\newblock On the rate of convergence in wasserstein distance of the empirical measure.
\newblock \emph{Probability theory and related fields}, 162\penalty0 (3):\penalty0 707--738, 2015.

\bibitem[Gaucher et~al.(2023)Gaucher, Schreuder, and Chzhen]{gaucher2023fair}
Solenne Gaucher, Nicolas Schreuder, and Evgenii Chzhen.
\newblock Fair learning with wasserstein barycenters for non-decomposable performance measures.
\newblock In \emph{International Conference on Artificial Intelligence and Statistics}, pp.\  2436--2459. PMLR, 2023.

\bibitem[Goh et~al.(2016)Goh, Cotter, Gupta, and Friedlander]{goh2016satisfying}
Gabriel Goh, Andrew Cotter, Maya Gupta, and Michael~P Friedlander.
\newblock Satisfying real-world goals with dataset constraints.
\newblock \emph{Advances in neural information processing systems}, 29, 2016.

\bibitem[Gong et~al.(2024)Gong, Usman, Zhao, and Inouye]{gong2024towards}
Ziyu Gong, Ben Usman, Han Zhao, and David~I Inouye.
\newblock Towards practical non-adversarial distribution matching.
\newblock In \emph{International Conference on Artificial Intelligence and Statistics}, pp.\  4276--4284. PMLR, 2024.

\bibitem[Gouic et~al.(2020)Gouic, Loubes, and Rigollet]{gouic2020projection}
Thibaut~Le Gouic, Jean-Michel Loubes, and Philippe Rigollet.
\newblock Projection to fairness in statistical learning.
\newblock \emph{arXiv preprint arXiv:2005.11720}, 2020.

\bibitem[Han et~al.(2023)Han, Lee, Wu, Kim, Wu, Wang, Xie, and Cha]{han2023dualfair}
Sungwon Han, Seungeon Lee, Fangzhao Wu, Sundong Kim, Chuhan Wu, Xiting Wang, Xing Xie, and Meeyoung Cha.
\newblock Dualfair: fair representation learning at both group and individual levels via contrastive self-supervision.
\newblock In \emph{Proceedings of the ACM web conference 2023}, pp.\  3766--3774, 2023.

\bibitem[Hardt et~al.(2016)Hardt, Price, and Srebro]{hardt2016equality}
Moritz Hardt, Eric Price, and Nati Srebro.
\newblock Equality of opportunity in supervised learning.
\newblock \emph{Advances in neural information processing systems}, 29, 2016.

\bibitem[He et~al.(2016)He, Zhang, Ren, and Sun]{he2016deep}
Kaiming He, Xiangyu Zhang, Shaoqing Ren, and Jian Sun.
\newblock Deep residual learning for image recognition.
\newblock In \emph{Proceedings of the IEEE conference on computer vision and pattern recognition}, pp.\  770--778, 2016.

\bibitem[Hong \& Yang(2021)Hong and Yang]{hong2021unbiased}
Youngkyu Hong and Eunho Yang.
\newblock Unbiased classification through bias-contrastive and bias-balanced learning.
\newblock \emph{Advances in Neural Information Processing Systems}, 34:\penalty0 26449--26461, 2021.

\bibitem[Hu et~al.(2023)Hu, Ratz, and Charpentier]{hu2023fairness}
Fran{\c{c}}ois Hu, Philipp Ratz, and Arthur Charpentier.
\newblock Fairness in multi-task learning via wasserstein barycenters.
\newblock \emph{arXiv preprint arXiv:2306.10155}, 2023.

\bibitem[Huang et~al.(2017)Huang, Liu, Van Der~Maaten, and Weinberger]{huang2017densely}
Gao Huang, Zhuang Liu, Laurens Van Der~Maaten, and Kilian~Q Weinberger.
\newblock Densely connected convolutional networks.
\newblock In \emph{Proceedings of the IEEE conference on computer vision and pattern recognition}, pp.\  4700--4708, 2017.

\bibitem[Irvin et~al.(2019)Irvin, Rajpurkar, Ko, Yu, Ciurea-Ilcus, Chute, Marklund, Haghgoo, Ball, Shpanskaya, et~al.]{irvin2019chexpert}
Jeremy Irvin, Pranav Rajpurkar, Michael Ko, Yifan Yu, Silviana Ciurea-Ilcus, Chris Chute, Henrik Marklund, Behzad Haghgoo, Robyn Ball, Katie Shpanskaya, et~al.
\newblock Chexpert: A large chest radiograph dataset with uncertainty labels and expert comparison.
\newblock In \emph{Proceedings of the AAAI conference on artificial intelligence}, volume~33, pp.\  590--597, 2019.

\bibitem[Jaschik(2023)]{IHE}
Scott Jaschik.
\newblock {Admissions Offices, Cautiously, Start Using AI}.
\newblock \url{https://www.insidehighered.com/news/admissions/2023/05/15/admissions-offices-cautiously-start-using-ai}, 2023.
\newblock Accessed: 2023-05-15.

\bibitem[Jiang et~al.(2020)Jiang, Pacchiano, Stepleton, Jiang, and Chiappa]{jiang2020wasserstein}
Ray Jiang, Aldo Pacchiano, Tom Stepleton, Heinrich Jiang, and Silvia Chiappa.
\newblock Wasserstein fair classification.
\newblock In \emph{Uncertainty in artificial intelligence}, pp.\  862--872. PMLR, 2020.

\bibitem[Kamiran \& Calders(2012)Kamiran and Calders]{kamiran2012data}
Faisal Kamiran and Toon Calders.
\newblock Data preprocessing techniques for classification without discrimination.
\newblock \emph{Knowledge and information systems}, 33\penalty0 (1):\penalty0 1--33, 2012.

\bibitem[Kim et~al.(2019)Kim, Kim, Kim, Kim, and Kim]{kim2019learning}
Byungju Kim, Hyunwoo Kim, Kyungsu Kim, Sungjin Kim, and Junmo Kim.
\newblock Learning not to learn: Training deep neural networks with biased data.
\newblock In \emph{Proceedings of the IEEE/CVF conference on computer vision and pattern recognition}, pp.\  9012--9020, 2019.

\bibitem[Kwegyir-Aggrey et~al.(2023)Kwegyir-Aggrey, Dai, Cooper, Dickerson, Hines, and Venkatasubramanian]{kwegyir2023repairing}
Kweku Kwegyir-Aggrey, Jessica Dai, A~Feder Cooper, John Dickerson, Keegan Hines, and Suresh Venkatasubramanian.
\newblock Repairing regressors for fair binary classification at any decision threshold.
\newblock In \emph{NeurIPS 2023 Workshop Optimal Transport and Machine Learning}, 2023.

\bibitem[Lindheim(2023)]{lindheim2023simple}
Johannes~von Lindheim.
\newblock Simple approximative algorithms for free-support wasserstein barycenters.
\newblock \emph{Computational Optimization and Applications}, 85\penalty0 (1):\penalty0 213--246, 2023.

\bibitem[Lipton et~al.(2018)Lipton, McAuley, and Chouldechova]{lipton2018does}
Zachary Lipton, Julian McAuley, and Alexandra Chouldechova.
\newblock Does mitigating ml's impact disparity require treatment disparity?
\newblock \emph{Advances in neural information processing systems}, 31, 2018.

\bibitem[Liu et~al.(2023)Liu, Wang, Wang, Wang, Su, and Gao]{liu2023simfair}
Tianci Liu, Haoyu Wang, Yaqing Wang, Xiaoqian Wang, Lu~Su, and Jing Gao.
\newblock Simfair: a unified framework for fairness-aware multi-label classification.
\newblock In \emph{Proceedings of the AAAI Conference on Artificial Intelligence}, volume~37, pp.\  14338--14346, 2023.

\bibitem[Liu et~al.(2018)Liu, Xu, Tsang, and Zhang]{liu2018metric}
Weiwei Liu, Donna Xu, Ivor~W Tsang, and Wenjie Zhang.
\newblock Metric learning for multi-output tasks.
\newblock \emph{IEEE Transactions on Pattern Analysis and Machine Intelligence}, 41\penalty0 (2):\penalty0 408--422, 2018.

\bibitem[Liu et~al.(2015)Liu, Luo, Wang, and Tang]{liu2015faceattributes}
Ziwei Liu, Ping Luo, Xiaogang Wang, and Xiaoou Tang.
\newblock Deep learning face attributes in the wild.
\newblock In \emph{Proceedings of International Conference on Computer Vision (ICCV)}, December 2015.

\bibitem[Lohia et~al.(2019)Lohia, Ramamurthy, Bhide, Saha, Varshney, and Puri]{lohia2019bias}
Pranay~K Lohia, Karthikeyan~Natesan Ramamurthy, Manish Bhide, Diptikalyan Saha, Kush~R Varshney, and Ruchir Puri.
\newblock Bias mitigation post-processing for individual and group fairness.
\newblock In \emph{Icassp 2019-2019 ieee international conference on acoustics, speech and signal processing (icassp)}, pp.\  2847--2851. IEEE, 2019.

\bibitem[Louizos et~al.(2015)Louizos, Swersky, Li, Welling, and Zemel]{louizos2015variational}
Christos Louizos, Kevin Swersky, Yujia Li, Max Welling, and Richard Zemel.
\newblock The variational fair autoencoder.
\newblock \emph{arXiv preprint arXiv:1511.00830}, 2015.

\bibitem[Madras et~al.(2018)Madras, Creager, Pitassi, and Zemel]{madras2018learning}
David Madras, Elliot Creager, Toniann Pitassi, and Richard Zemel.
\newblock Learning adversarially fair and transferable representations.
\newblock In \emph{International Conference on Machine Learning}, pp.\  3384--3393. PMLR, 2018.

\bibitem[Mo et~al.(2021)Mo, Kang, Sohn, Li, and Shin]{mo2021object}
Sangwoo Mo, Hyunwoo Kang, Kihyuk Sohn, Chun-Liang Li, and Jinwoo Shin.
\newblock Object-aware contrastive learning for debiased scene representation.
\newblock \emph{Advances in Neural Information Processing Systems}, 34:\penalty0 12251--12264, 2021.

\bibitem[Nadaraya(1964)]{nadaraya1964estimating}
Elizbar~A Nadaraya.
\newblock On estimating regression.
\newblock \emph{Theory of Probability \& Its Applications}, 9\penalty0 (1):\penalty0 141--142, 1964.

\bibitem[O'neil(2017)]{o2017weapons}
Cathy O'neil.
\newblock \emph{Weapons of math destruction: How big data increases inequality and threatens democracy}.
\newblock Crown, 2017.

\bibitem[Park et~al.(2022)Park, Lee, Lee, Hwang, Kim, and Byun]{park2022fair}
Sungho Park, Jewook Lee, Pilhyeon Lee, Sunhee Hwang, Dohyung Kim, and Hyeran Byun.
\newblock Fair contrastive learning for facial attribute classification.
\newblock In \emph{Proceedings of the IEEE/CVF Conference on Computer Vision and Pattern Recognition}, pp.\  10389--10398, 2022.

\bibitem[Petersen et~al.(2021)Petersen, Mukherjee, Sun, and Yurochkin]{petersen2021post}
Felix Petersen, Debarghya Mukherjee, Yuekai Sun, and Mikhail Yurochkin.
\newblock Post-processing for individual fairness.
\newblock \emph{Advances in Neural Information Processing Systems}, 34:\penalty0 25944--25955, 2021.

\bibitem[Peyr{\'e} et~al.(2017)Peyr{\'e}, Cuturi, et~al.]{peyre2017computational}
Gabriel Peyr{\'e}, Marco Cuturi, et~al.
\newblock Computational optimal transport.
\newblock \emph{Center for Research in Economics and Statistics Working Papers}, \penalty0 (2017-86), 2017.

\bibitem[Qi et~al.(2024)Qi, Hu, Lin, and Yang]{qi2024provable}
Qi~Qi, Quanqi Hu, Qihang Lin, and Tianbao Yang.
\newblock Provable optimization for adversarial fair self-supervised contrastive learning.
\newblock \emph{arXiv preprint arXiv:2406.05686}, 2024.

\bibitem[Radford et~al.(2021)Radford, Kim, Hallacy, Ramesh, Goh, Agarwal, Sastry, Askell, Mishkin, Clark, et~al.]{radford2021learning}
Alec Radford, Jong~Wook Kim, Chris Hallacy, Aditya Ramesh, Gabriel Goh, Sandhini Agarwal, Girish Sastry, Amanda Askell, Pamela Mishkin, Jack Clark, et~al.
\newblock Learning transferable visual models from natural language supervision.
\newblock In \emph{International conference on machine learning}, pp.\  8748--8763. PMLR, 2021.

\bibitem[Raji \& Buolamwini(2019)Raji and Buolamwini]{raji2019actionable}
Inioluwa~Deborah Raji and Joy Buolamwini.
\newblock Actionable auditing: Investigating the impact of publicly naming biased performance results of commercial ai products.
\newblock In \emph{Proceedings of the 2019 AAAI/ACM Conference on AI, Ethics, and Society}, pp.\  429--435, 2019.

\bibitem[Ramaswamy et~al.(2021)Ramaswamy, Kim, and Russakovsky]{ramaswamy2021fair}
Vikram~V Ramaswamy, Sunnie~SY Kim, and Olga Russakovsky.
\newblock Fair attribute classification through latent space de-biasing.
\newblock In \emph{Proceedings of the IEEE/CVF conference on computer vision and pattern recognition}, pp.\  9301--9310, 2021.

\bibitem[Santambrogio(2015)]{santambrogio2015optimal}
Filippo Santambrogio.
\newblock Optimal transport for applied mathematicians.
\newblock \emph{Birk{\"a}user, NY}, 55\penalty0 (58-63):\penalty0 94, 2015.

\bibitem[Sarhan et~al.(2020)Sarhan, Navab, Eslami, and Albarqouni]{sarhan2020fairness}
Mhd~Hasan Sarhan, Nassir Navab, Abouzar Eslami, and Shadi Albarqouni.
\newblock Fairness by learning orthogonal disentangled representations.
\newblock In \emph{Computer Vision--ECCV 2020: 16th European Conference, Glasgow, UK, August 23--28, 2020, Proceedings, Part XXIX 16}, pp.\  746--761. Springer, 2020.

\bibitem[Schreuder \& Chzhen(2021)Schreuder and Chzhen]{schreuder2021classification}
Nicolas Schreuder and Evgenii Chzhen.
\newblock Classification with abstention but without disparities.
\newblock In \emph{Uncertainty in Artificial Intelligence}, pp.\  1227--1236. PMLR, 2021.

\bibitem[Simonite(2015)]{simonite2015probing}
Tom Simonite.
\newblock Probing the dark side of google’s ad-targeting system.
\newblock \emph{MIT Technology Review}, 2015.

\bibitem[Tifrea et~al.(2024)Tifrea, Lahoti, Packer, Halpern, Beirami, and Prost]{tifreafrappe}
Alexandru Tifrea, Preethi Lahoti, Ben Packer, Yoni Halpern, Ahmad Beirami, and Flavien Prost.
\newblock Frapp{\'e}: A group fairness framework for post-processing everything.
\newblock In \emph{Forty-first International Conference on Machine Learning}, 2024.

\bibitem[Vogel et~al.(2021)Vogel, Bellet, and Cl{\'e}men{\c{c}}on]{vogel2021learning}
Robin Vogel, Aur{\'e}lien Bellet, and Stephan Cl{\'e}men{\c{c}}on.
\newblock Learning fair scoring functions: Bipartite ranking under roc-based fairness constraints.
\newblock In \emph{International conference on artificial intelligence and statistics}, pp.\  784--792. PMLR, 2021.

\bibitem[Xian et~al.(2023)Xian, Yin, and Zhao]{xian2023fair}
Ruicheng Xian, Lang Yin, and Han Zhao.
\newblock Fair and optimal classification via post-processing.
\newblock In \emph{International Conference on Machine Learning}, pp.\  37977--38012. PMLR, 2023.

\bibitem[Xie et~al.(2017)Xie, Dai, Du, Hovy, and Neubig]{xie2017controllable}
Qizhe Xie, Zihang Dai, Yulun Du, Eduard Hovy, and Graham Neubig.
\newblock Controllable invariance through adversarial feature learning.
\newblock \emph{Advances in neural information processing systems}, 30, 2017.

\bibitem[Xu et~al.(2019)Xu, Shi, Tsang, Ong, Gong, and Shen]{xu2019survey}
Donna Xu, Yaxin Shi, Ivor~W Tsang, Yew-Soon Ong, Chen Gong, and Xiaobo Shen.
\newblock Survey on multi-output learning.
\newblock \emph{IEEE transactions on neural networks and learning systems}, 31\penalty0 (7):\penalty0 2409--2429, 2019.

\bibitem[Yang et~al.(2023)Yang, Ko, Varshney, and Ying]{yang2023minimax}
Zhenhuan Yang, Yan~Lok Ko, Kush~R Varshney, and Yiming Ying.
\newblock Minimax auc fairness: Efficient algorithm with provable convergence.
\newblock In \emph{Proceedings of the AAAI Conference on Artificial Intelligence}, volume~37, pp.\  11909--11917, 2023.

\bibitem[Yao et~al.(2023)Yao, Lin, and Yang]{yao2023stochastic}
Yao Yao, Qihang Lin, and Tianbao Yang.
\newblock Stochastic methods for auc optimization subject to auc-based fairness constraints.
\newblock In \emph{International Conference on Artificial Intelligence and Statistics}, pp.\  10324--10342. PMLR, 2023.

\bibitem[Yuan et~al.(2022)Yuan, Wu, Qiu, Du, Zhang, Zhou, and Yang]{yuan2022provable}
Zhuoning Yuan, Yuexin Wu, Zi-Hao Qiu, Xianzhi Du, Lijun Zhang, Denny Zhou, and Tianbao Yang.
\newblock Provable stochastic optimization for global contrastive learning: Small batch does not harm performance.
\newblock In \emph{International Conference on Machine Learning}, pp.\  25760--25782. PMLR, 2022.

\bibitem[Zeng et~al.(2022)Zeng, Dobriban, and Cheng]{zeng2022bayes}
Xianli Zeng, Edgar Dobriban, and Guang Cheng.
\newblock Bayes-optimal classifiers under group fairness.
\newblock \emph{arXiv preprint arXiv:2202.09724}, 2022.

\bibitem[Zhang et~al.(2018)Zhang, Lemoine, and Mitchell]{zhang2018mitigating}
Brian~Hu Zhang, Blake Lemoine, and Margaret Mitchell.
\newblock Mitigating unwanted biases with adversarial learning.
\newblock In \emph{Proceedings of the 2018 AAAI/ACM Conference on AI, Ethics, and Society}, pp.\  335--340, 2018.

\bibitem[Zhang et~al.(2022)Zhang, Kuang, Chen, Liu, Wu, and Xiao]{zhang2022fairness}
Fengda Zhang, Kun Kuang, Long Chen, Yuxuan Liu, Chao Wu, and Jun Xiao.
\newblock Fairness-aware contrastive learning with partially annotated sensitive attributes.
\newblock In \emph{The Eleventh International Conference on Learning Representations}, 2022.

\bibitem[Zhang et~al.(2017)Zhang, Song, and Qi]{zhifei2017cvpr}
Zhifei Zhang, Yang Song, and Hairong Qi.
\newblock Age progression/regression by conditional adversarial autoencoder.
\newblock In \emph{IEEE Conference on Computer Vision and Pattern Recognition (CVPR)}. IEEE, 2017.

\end{thebibliography}
\bibliographystyle{tmlr}

\newpage
\appendix

\section{Proofs}

Before providing the proof of Theorem~\ref{thm:char}, let's review some additional results about Optimal Transport theory.

The next result shows that as long as two measures admit a density with finite second moments, there exists a unique deterministic optimal transport map between them.
\begin{lemma}
(Theorem 1.22 in \citet{santambrogio2015optimal}) Let $\mu, \nu$ be two measures on $\R^k$ with finite second moments such that $\mu$ has a density and let $X \sim \mu$. Then there exists a unique deterministic mapping $T: \R^k \rightarrow \R^k$ such that
\[
\cW_2^2(\mu, \nu) = \E\|X-T(X)\|^2_2,
\]
that is $(X, T(X)) \sim \bar\gamma \in \Gamma_{\mu, \nu}$ where $\bar \gamma$ is an optimal coupling.
\label{lem:map}
\end{lemma}

\subsection{Proof of Theorem~\ref{thm:char}}
This proof is originally from the proof of Theorem 2.3 in \citet{chzhen2020fair}. We extend their results from single output case to the multi-output case with some modifications in their proofs.

\textbf{Theorem~\ref{thm:char}}
Suppose $\nu_{f^*|s}$ has density and finite second moments for each $s \in \cS$. Then
\begin{small}
\begin{align*}
\min_{\mathcal{U}(f)=0}\mathcal{R}(f) = \mathcal{U}(f^*)=\min_\nu \sum_{s \in \cS} p_s \cW^2_2(\nu_{f^*|s}, \nu).
\end{align*}    
\end{small}
Moreover, if $f_0$ and $\nu_0$ solve the first and second minimization in \eqref{eqn:char}, respectively, then $\nu_0$ is the density of $f_0$ and
\begin{small}
\begin{align*}
f_0(x,s) =T_{f^*|s,\nu^0}(f^*(x,s))
\end{align*}    
\end{small}  
where $T_{f^*|s,\nu_0}:\R^k \rightarrow \R^k$ is the optimal transport mapping from $\nu_{f^*|s}$ to $\nu_0$.

\begin{proof}
First, we'd like to show that 
\begin{align*}
\min_{\mathcal{U}(f)=0} \E\|f^*(X,S)-f(X,S)\|^2 \geq \min_\nu \sum_{s \in S} p_s \cW^2_2(\nu_{f^*|s}, \nu).
\end{align*}
Let $\bar g: \R^{d} \times \cS \rightarrow \R^k$ be the minimizer of the l.h.s of the above equation and denoted by $\nu_{\bar g}$ the distribution of $\bar g$. Since $\nu_{f^*|s}$ admits density, with Lemma \ref{lem:map}, for each $s\in S$ there exists $T_{f*|s, \bar g}$ such that
\begin{align*}
    \sum_{s\in S}p_s \cW_2^2(\nu_{f^*|s},\nu_{\bar g}) &= \sum_{s\in S}p_s\int \|z- T_{f^*|s,\bar g}(z)\|^2 d\nu_{f^*|s}(z) \\
    &= \sum_{s\in S}p_s \int_{\R^d}\|f^*(x,s) - T_{f^*|s,\bar g}(f^*(x,s))\|^2 d\P_{X|S=s}(x) \\
    &= \sum_{s\in S}p_s\E\big[ \|f^*(X,s) - T_{f^*|s,\bar g}(f^*(X,s))\|^2 | S=s\big] \\
    &= \E\|f^*(X,S) - \tilde g(X,S)\|^2
\end{align*}
where we define  $\tilde g(x,s) = T_{f^*|s,\bar g}(f^*(x,s)) $  for all $(x,s)\in \R^d \times S$. With optimal transportation, $\tilde g(X,s)|{S=s}$ follow the distribution  $\nu_{\bar g}$ for any
$s \in S$. Then we have
\begin{align*}
\cU(\tilde g) =\min_{\nu}\sum_{s\in \cS} p_s\cW^2_2(\nu_{\tilde g|s},\nu) =0 
\end{align*}     
which indicates $\tilde g$ is fair. By optimality of $\bar g$ we have
\begin{align*}
    \E\|f^*(X,S) - \tilde g(X,S)\|^2 \geq \E\|f^*(X,S) - \bar g(X,S)\|^2
\end{align*}
Due to definition of $\cW_2^2$, for each $s\in S$ we have
\begin{align*}
    \cW_2^2(\nu_{f^*|s},\nu_{\bar g}) \leq \E[\|f^*(X,S) - \bar g(X,S)\|^2 | S=s]
\end{align*}
Then we can conclude
\begin{align*}
     \sum_{s \in S} p_s \cW^2_2(\nu_{f^*|s}, \nu_{\bar g}) = \min_{\mathcal{U}(f)=0} \E\|f^*(X,S)-f(X,S)\|^2
\end{align*}
This implies that
\begin{align}
\min_{\mathcal{U}(f)=0} \E\|f^*(X,S)-f(X,S)\|^2 \geq \min_\nu \sum_{s \in S} p_s \cW^2_2(\nu_{f^*|s}, \nu).
\label{eqn:pos}
\end{align}
Second, we are going to show that the opposite inequality also holds. To this end, we define $\nu_0$ as 
\begin{align*}
    \nu_0 \in \arg \min_\nu \sum_{s\in S}p_s \cW_2^2(\nu_{f^*|s}, \nu)
\end{align*}
Since we assume $\nu_{f^*|s}$ admits density, with Lemma \ref{lem:map}, there exists  $T_{\nu_{f^*|s}, \nu_0}$ as a optimal transport map from $\nu_{f^*|s}$ to $ \nu_0$. And we define $f_0$ for all $(x,s) \in \R^d \times S$ as
\begin{align*}
    f_0(x,s) = T_{\nu_{f^*|s}, \nu_0} \circ f_0(x,s)
\end{align*}
By the definition of $f_0$ and the Lemma \ref{lem:map}, we have
\begin{align}
    \min_{\nu} \sum_{s \in S} p_s \cW^2_2(\nu_{f^*|s}, \nu) = \E\|f^*(X,S)-f_0(X,S)\|^2
\label{eqn:eq}
\end{align}
Moreover since $\nu_0$ is independent from $S$, using similar argument as above we can show that $f_0$ is fair, and it yields
\begin{align}
\min_\nu \sum_{s \in S} p_s \cW^2_2(\nu_{f^*|s}, \nu) \geq \min_{\mathcal{U}(f)=0} \E\|f^*(X,S)-f(X,S)\|^2.
\label{eqn:opp}
\end{align}
Therefore, combining Eq.~\ref{eqn:pos} and Eq.~\ref{eqn:opp}, we showed that 
\begin{align*}
    \min_\nu \sum_{s \in S} p_s \cW^2_2(\nu_{f^*|s}, \nu) = \min_{\mathcal{U}(f)=0} \E\|f^*(X,S)-f(X,S)\|^2. 
\end{align*}
Thanks to Eq.~\ref{eqn:eq}, we can also have
\begin{align*}
    \E\|f^*(X,S)-f_0(X,S)\|^2 = \E\|f^*(X,S)-\bar g(X,S)\|^2
\end{align*}
and since $f_0$ is fair we can put $\bar g = f_0$. This proof is concluded. 
\end{proof}

\subsection{Proof of Proposition~\ref{lem:kernel_map}}
\label{sec:prooflemma}
\textbf{Proposition~\ref{lem:kernel_map}}
Suppose $K(z)=\frac{1}{\sqrt{2\pi}}\exp(-z^2)$. Conditioning on data $D=\{(x_i, s_i)\}_{i=1}^n$, it holds that
$$
\cU(\bar f(X,S))\leq \sum_{s\in\cS}O\left(p_sn_s^2\exp\left(-\frac{1}{h^2}\right)\right)+\sum_{s\in\cS}O\left(\frac{p_s}{h^4}\cW^2_2(\nu_{f^*|s},\nu_{f^*(D_s)})\right).
$$
\begin{proof}
{
Recall that $(X,S)$ is the ground-truth distribution. Let 
$$
\bar f(x,s):=\tilde T_{\nu_{f^*(D_s)}, \tilde \nu_0}(f^*(x,s)).
$$
Let $\nu_{\bar f}$ be the distribution of $\bar f(X,S)$, and $\nu_{\bar f|s}$ be the distribution of $\bar f(X,S)$ conditioning on $S=s$ for $s\in\cS$. Suppose $\tilde \nu_0$ has $n_0$ supports, denoted by $\{\xi_i\}_{i=1}^{n_0}\subset\mathbb{R}^k$. In the entire proof, we assume $D_s$ is already sampled and thus deterministic for $s\in\cS$.
}

{
By \eqref{eq:Uf}, we have
\begin{equation}
\label{eq:Ufreplace}
\begin{aligned}
\mathcal{U}(\bar f) =\min_{\nu}\sum_{s\in \cS} p_s\cW^2_2(\nu_{\bar f|s},\nu)
&\leq \sum_{s\in \cS} p_s\cW^2_2(\nu_{\bar f|s},\tilde\nu_0) \\
&\leq \sum_{s\in \cS} 2p_s\cW^2_2(\nu_{\bar f|s},\nu_{\bar f(D_s)})
+\sum_{s\in \cS} 2p_s\cW^2_2(\nu_{\bar f(D_s)},\tilde\nu_0),
\end{aligned}
\end{equation}
where the second inequality is the triangle inequality. Next, we bound $\cW^2_2(\nu_{\bar f|s},\nu_{\bar f(D_s)})$ and $\cW^2_2(\nu_{\bar f(D_s)},\tilde\nu_0)$ from above separately.}

{
Consider any $s\in\cS$. We denote the discrete distribution of $\bar f(X,S)$ when $(X,S)$ is uniformly sampled from $D_s$ as $\nu_{\bar f(D_s)}$.
Let $\phi(\cdot):\mathbb{R}^k\rightarrow\mathbb{R}$ be any $1$-Lipschitz continuous function on $\mathbb{R}^k$, meaning that 
$|\phi(\xi)-\phi(\xi')|\leq\|\xi-\xi'\|_2$ for any $\xi$ and $\xi'$ in $\mathbb{R}^k$. We have 
$$
\int_{\mathbb{R}^k} \phi(\xi)d\nu_{\bar f(D_s)}=\frac{1}{n_s}\sum_{i=1}^{n_s}\mathbb{E}\phi(\tilde T_{\nu_{f^*(D_s)}, \tilde \nu_0}(f^*(x_i^s,s))\text{ and }\int_{\mathbb{R}^k} \phi(\xi)d\tilde \nu_0=\frac{1}{n_s}\sum_{i=1}^{n_s}\mathbb{E}\phi(T_{\nu_{f^*(D_s)}, \tilde \nu_0}(f^*(x_i^s,s))).
$$
Consider any $i\in\{1,\dots,n_s\}$. Let $\mathbb{E}_i$ be the conditional expectation conditioning on the outcome of random mapping $ T_{\nu_{f^*(D_s)}, \tilde \nu_0}(f^*(x_i^s,s)$.
By \eqref{eq:test_T} and the 1-Lipschitz continuity of $\phi(\cdot)$, we have 
\small
\begin{align}
\nonumber
&\left|\mathbb{E}\phi(\tilde T_{\nu_{f^*(D_s)}, \tilde \nu_0}(f^*(x_i^s,s))-\mathbb{E}\phi(T_{\nu_{f^*(D_s)}, \tilde \nu_0}(f^*(x_i^s,s)))\right|\\\nonumber
\leq&\left|\mathbb{E}\left[\mathbb{E}_i\phi(\tilde T_{\nu_{f^*(D_s)}, \tilde \nu_0}(f^*(x_i^s,s))-\mathbb{E}_i\phi(T_{\nu_{f^*(D_s)}, \tilde \nu_0}(f^*(x_i^s,s)))\right]\right|\\\nonumber
\leq&\frac{\sum_{j=1}^{n_s}K\left((f^*(x_i^s,s)-f^*(x_j^s,s))/h\right) \mathbb{E}\mathbb{E}_i\left|T_{\nu_{f^*(D_s)}, \tilde \nu_0}(f^*(x_i^s,s))-T_{\nu_{f^*(D_s)}, \tilde \nu_0}(f^*(x_j^s,s))\right|
}{\sum_{j=1}^{n_s} K\left((f^*(x_i^s,s)-f^*(x_j^s,s))/h\right)}\\\nonumber
=&\frac{\sum_{j=1,j\neq i}^{n_s}K\left((f^*(x_i^s,s)-f^*(x_j^s,s))/h\right) \mathbb{E}\mathbb{E}_i\left|T_{\nu_{f^*(D_s)}, \tilde \nu_0}(f^*(x_i^s,s))-T_{\nu_{f^*(D_s)}, \tilde \nu_0}(f^*(x_j^s,s))\right|
}{\sum_{j=1}^{n_s} K\left((f^*(x_i^s,s)-f^*(x_j^s,s))/h\right)}\\\nonumber
\leq&n_s\exp\left(-\frac{\Delta_{f^*(D_s),\min}^2}{2h^2}\right)\Delta_{\tilde \nu_0,\max},
\end{align}
\normalsize
where 
$$
\Delta_{f^*(D_s),\min}:=\min_{i,j=1,\dots,n_s, i\neq j}\|f^*(x_i^s,s)-f^*(x_j^s,s)\|\text{  and  }
\Delta_{\tilde \nu_0,\max}:=\max_{j,j'=1,\dots,n_0}\|\xi_j-\xi_{j'}\|.
$$
As a result, we have
$$
\int_{\mathbb{R}^k} \phi(\xi)d\nu_{\bar f(D_s)}-\int_{\mathbb{R}^k} \phi(\xi)d\tilde \nu_0\leq
n_s\exp\left(-\frac{\Delta_{f^*(D_s),\min}^2}{2h^2}\right)\Delta_{\tilde \nu_0,\max}
$$
for any $\phi$ that is $1$-Lipschitz continuous function on $\mathbb{R}^k$. By Kantorovich-Rubinstein duality, we have 
\begin{equation}
\label{eq:ineq1}
\cW^2_2(\nu_{\bar f(D_s)},\tilde \nu_0)\leq n_s^2\exp\left(-\frac{\Delta_{f^*(D_s),\min}^2}{h^2}\right)\Delta_{\tilde \nu_0,\max}^2=O\left(n_s^2\exp\left(-\frac{1}{h^2}\right)\right).
\end{equation}
}



{
Again, let $\phi(\cdot):\mathbb{R}^k\rightarrow\mathbb{R}$ be any $1$-Lipschitz continuous function on $\mathbb{R}^k$. 
Let $\mathbb{E}_T$ be the expectation over the random mappings $T_{\nu_{f^*(D_s)}, \tilde \nu_0}(f^*(x_i^s,s))$ for $i=1,\dots,n_s$. 
We have 
$$
\int_{\mathbb{R}^k} \phi(\xi)d\nu_{\bar f(D_s)}=\int_{\mathbb{R}^k} \mathbb{E}_T\Phi(\xi)d\nu_{f^*(D_s)}\text{ and }\int_{\mathbb{R}^k} \phi(\xi)d\nu_{\bar f|s}=\int_{\mathbb{R}^k} \mathbb{E}_T\Phi(\xi)d\nu_{f^*|s},
$$
where
$$
\Phi(\xi)=\phi\left(\frac{\sum_{i=1}^{n_s}K\left(\frac{\xi-f^*(x_i^s,s)}{h}\right)T_{\nu_{f^*(D_s)}, \tilde \nu_0}(f^*(x_i^s,s))
}{\sum_{i=1}^{n_s} K\left(\frac{\xi-f^*(x_i^s,s)}{h}\right)}\right)
$$
is a random value function whose randomness is from $T_{\nu_{f^*(D_s)}, \tilde \nu_0}(f^*(x_i^s,s))$ for $i=1,\dots,n_s$. Since $K$ is $O(\frac{1}{h^2})$-Lipschitz continuous and $\phi$ is 1-Lipschitz continuous, $\Phi(\xi)$ is  $O(\frac{1}{h^2})$-Lipschitz continuous in $\xi$ given any outcomes of random mappings $T_{\nu_{f^*(D_s)}, \tilde \nu_0}(f^*(x_i^s,s))$ for $i=1,\dots,n_s$. Therefore, $\mathbb{E}_T\Phi(\xi)$ is also $O(\frac{1}{h})$-Lipschitz continuous. 
}

{
By Kantorovich-Rubinstein inequality~\citep{edwards2011kantorovich}, there exists a constant $C$ such that 
\begin{equation*}
\int_{\mathbb{R}^k} \phi(\xi)d\nu_{\bar f(D_s)}-\int_{\mathbb{R}^k} \phi(\xi)d\nu_{\bar f|s}\leq 
\int_{\mathbb{R}^k} \mathbb{E}_T\Phi(\xi)d\nu_{f^*(D_s)}-\int_{\mathbb{R}^k} \mathbb{E}_T\Phi(\xi)d\nu_{f^*|s}\leq
\frac{C}{h^2}\cW_2(\nu_{f^*|s},\nu_{f^*(D_s)}).
\end{equation*}
which implies
\begin{equation}
\label{eq:ineq2}
\cW^2_2(\nu_{\bar f(D_s)},\tilde\nu_0)\leq
\frac{C^2}{h^4}\cW^2_2(\nu_{f^*|s},\nu_{f^*(D_s)}).
\end{equation}
Applying \eqref{eq:ineq1} and \eqref{eq:ineq2} to \eqref{eq:Ufreplace} gives the first conclusion.
}

{
Consider any two groups $s$ and $s'$ in $\cS$. Applying \eqref{eq:ineq1} and \eqref{eq:ineq2} and the triangle inequality again, we have 
\begin{align*}
\cW^2_2(\nu_{\bar f|s},\nu_{\bar f|s'})\leq&\cW^2_2(\nu_{\bar f|s},\tilde\nu_0)+\cW^2_2(\nu_{\bar f|s'},\tilde\nu_0)\\
\leq&\cW^2_2(\nu_{\bar f|s},\nu_{\bar f(D_s)})+\cW^2_2(\nu_{\bar f(D_s)},\tilde\nu_0)
+\cW^2_2(\nu_{\bar f|s'},\tilde\nu_0)+\cW^2_2(\nu_{\bar f(D_{s'})},\tilde\nu_0)\\
\leq & O\left((n_s^2+n_{s'}^2)\exp\left(-\frac{1}{h^2}\right)\right)+ O\left(\frac{1}{h^4}\cW^2_2(\nu_{f^*|s},\nu_{f^*(D_s)})\right),
\end{align*}
which gives the second conclusion.
}
\end{proof}



\section{Details of Datasets}
\label{sec:data}
In this section, we provide more details on the datasets we used in the numerical experiments. We include four datasets from various domains, including marketing domain(\textit{Customer} dataset \footnote{\url{https://www.kaggle.com/datasets/kaushiksuresh147/customer-segmentation}}), medical diagnosis( \textit{Chexpert} Dataset~\citep{irvin2019chexpert}), face recognition (\textit{CelebA} dataset~\citep{liu2015faceattributes} and \textit{UTKFace} dataset~\citep{zhifei2017cvpr}).

The \textit{Customer} dataset has $8068$ training samples and $2627$ testing samples and the task is to classify customers into anonymous customer categories for target marketing. We partition the data into four sensitive groups based on the gender and the marital status of customers: married female, unmarried female, married male, and unmarried male.

The \textit{Chexpert}  dataset contains 224,316 training instances and the task is to detect five chest and lung diseases based on X-ray images. Due to the high computational complexity of solving optimal transportation between large datasets, we sample $5\%$ instances from the original training data as the training set and sample another $5\%$ as the testing set. The \textit{CelebA}  dataset contains 162,770 training instances and 39,829 testing instances and the task is to detect ten attributes (chosen based on \citet{ramaswamy2021fair}) of the person in an image, which are being attractive, having bags under eye, having black hair, having bangs, wearing glasses, having high cheek bones, being smiling, wearing hat, having a slightly open mouth, and have a pointy nose. For the same computational reason, we sample $5\%$ instances from the original training data as the training set and sample $20\%$  from the original testing data as the testing set. For both Chexpert and CelebA datasets, we partition the data into four sensitive groups based on gender and age: young female, old female, young male, and old male. \textit{UTKFace} dataset consists of 23705 face images with five groups in terms of race(i.e., White, Black, Asian, Indian, and Others) and we randomly split it into training and testing (8:2) sets. And the task is to classify gender and age ($25\leq \text{age}\leq60$, customized by us) based on face images. All the data in \textit{UTKFace} dataset are utilized.

\end{document}